\RequirePackage{fix-cm}
\documentclass[twocolumn]{svjour3}          % twocolumn
\smartqed  % flush right qed marks, e.g. at end of proof
\usepackage{graphicx}
\usepackage{subfigure}
\usepackage{epsfig}
\usepackage{amsmath}
\usepackage{amssymb}
\usepackage{booktabs}
\usepackage{rotating}
\usepackage{multirow}
\usepackage{color}
\usepackage{xspace}
\usepackage{tabularx}

\usepackage{hhline}
\usepackage{adjustbox}

\usepackage{mathtools}
\usepackage{comment}

\usepackage{multirow}
\usepackage{enumitem}

\newcommand{\subparagraph}{}

\usepackage{titlesec}
\usepackage{bm}
\titleformat{\paragraph}
{\normalfont\normalsize\it}{\theparagraph}{0.5em}{}
\titlespacing*{\paragraph}
{0pt}{3.25ex plus 1ex minus .2ex}{1.5ex plus .2ex}

\DeclareMathOperator*{\argmin}{argmin}

\makeatletter
\DeclareRobustCommand\onedot{\futurelet\@let@token\@onedot}
\def\@onedot{\ifx\@let@token.\else.\null\fi\xspace}

\newcommand\footnoteref[1]{\protected@xdef\@thefnmark{\ref{#1}}\@footnotemark}

\makeatother

\begin{document}

\def\HS{\hspace{\fontdimen2\font}}

%\title{Generating faces for affect analysis}

\title{Deep Neural Network Augmentation:  Generating Faces for Affect Analysis}

\titlerunning{Generating faces for affect analysis}

\date{Received: Sept 30th 2018 / Accepted: date}

\author{Dimitrios Kollias $^\star$        \and
        Shiyang Cheng $^\dagger$\and
        Evangelos Ververas $^*$ \and 
        Irene Kotsia$^{1}$ \and 
        Stefanos Zafeiriou$^{2}$
}

\authorrunning{D. Kollias}

\institute{    
$^\star$dimitrios.kollias15@imperial.ac.uk\\
$^\dagger$shiyang.cheng11@imperial.ac.uk\\
$^*$e.ververas16@imperial.ac.uk\\
$^{1}$I.Kotsia@mdx.ac.uk\\
$^{2}$s.zafeiriou@imperial.ac.uk\\
\at
$^{\star,\dagger,*,2}$Department of Computing, Imperial College London, Queen’s Gate, London SW7 2AZ, UK\\
$^1$ Department of Computer Science, Middlesex University
of London, London NW4 4BT, U.K\\
%$^{2}$Center for Machine Vision and Signal Analysis, University of
%Oulu, Oulu, Finland\\
}

%% ERRORS: check tis images na einai oles apo ta swsta atoma;
%%% episis fig. 12 ta histograms stn affectnet den einai sto union of manually+automatically annotated alla mono automatically annotated

%%%% ws additional to synthesize basic expressions kane annotate ta VA mean faces se expressions k use auta

\maketitle

\begin{abstract}

This paper presents a novel approach for synthesizing facial affect; either in terms of the six basic expressions (i.e., anger, disgust, fear, joy, sadness and surprise), or in terms of valence (i.e., how positive or negative is an emotion) and arousal (i.e., power of the emotion activation). 
The proposed approach accepts the following inputs: i) a neutral 2D image of a person; ii) a basic facial expression or a pair of valence-arousal (VA) emotional state descriptors to be generated, or a path of affect in the 2D VA Space to be generated as an image sequence. In order to synthesize affect in terms of VA, for this person, $600,000$ frames from the 4DFAB database were annotated. The affect synthesis is implemented by fitting a 3D Morphable Model on the neutral image, then  deforming  the  reconstructed  face and adding the inputted affect, and  blending the new face with  the  given  affect  into  the  original  image.
Qualitative experiments illustrate the generation of realistic images, when the neutral image is sampled from thirteen well known lab-controlled or in-the-wild databases, including  Aff-Wild, AffectNet, RAF-DB; comparisons with Generative Adversarial Networks (GANs) show the higher quality achieved by the proposed approach. 
Then, quantitative experiments are conducted, in which the synthesized images are used for data augmentation in training Deep Neural Networks to perform affect recognition over all databases; greatly improved performances are achieved when compared with state-of-the-art methods, as well as with GAN-based data augmentation, in all cases.

  %add tis dbs pou ekanes ta experiments
\end{abstract}

\section{Introduction}

Rendering photorealistic facial expressions from single
static faces while preserving the identity information is an open research topic which has 
significant impact on the area of affective computing.
Generating faces of a specific person with different facial
expressions can be used in various applications, including face
recognition~\cite{cao2018vggface2}~\cite{parkhi2015deep}, face verification~\cite{sun2014deep}~\cite{taigman2014deepface}, emotion prediction, expression database generation, facial expression augmentation and entertainment. 

This paper describes a novel approach that uses an arbitrary face
image with a neutral
expression and synthesizes a new face image of the same
person, but with a different expression,  generated according to a categorical or dimensional emotion representation model. 
This problem cannot be tackled using small databases
with labeled facial expressions, as it would be really difficult to disentangle facial expressions and identity information through them.
Our approach is based on the analysis of a large 4D facial database, the 4DFAB \cite{cheng4dfab}, which we appropriately annotated and used for facial expression synthesis on a given subject's face.

At first, a dimensional emotion model, in terms of the continuous variables, valence (i.e., how positive or negative is an emotion) and arousal (i.e., power of the emotion activation) \cite{whissell1989dictionary} \cite{russell1978evidence}, has been used to annotate a large amount of 600,000 facial images.  This model can represent, not only primary, extreme expressions, but also subtle expressions which are met in everyday human to human, or human to machine interactions. Additionally, a categorical emotion model, in terms of the six basic facial expressions, has been used, according to which 12,000 expressions from the 4DFAB were selected, including 2,000 cases for each of the six basic expressions.

The proposed approach accepts: i) a pair of valence-arousal values and synthesize the respective facial affect, ii) a path of affect in the 2D VA Space and synthesize a temporal sequence showing it, iii) a value indicating the basic facial expression to be synthesized; a given neutral 2D image of a person is used in all cases to appropriately transfer the synthesized affect. 

Section 2 refers to related work regarding facial expression synthesis, as well as data augmentation related methodologies. Section 3 presents materials and methods that are used in the current work. We describe the annotation and use of the 4DFAB database and the 3D Morphable Model that we utilize in our developments. Section 4 presents our approach, explaining in detail all steps used to synthesize affect on an image or image sequence.
Section 5 mentions the categorical and dimensional databases, which are used by our approach. 

An extensive experimental study is presented in Section 6. At first, a qualitative evaluation of the proposed approach is provided, also showing the achieved higher quality when compared to GAN-generated facial affect. Then, we use the synthesized facial images for data augmentation and train Deep Neural Networks over eight databases, annotated with either dimensional or categorical affect labels. We show that the achieved performance is much higher than i) that obtained by the respective state-of-the-art methods, ii) the performance of the same DNNs with data augmentation provided by the StarGAN and Ganimation networks. A further comparison with GANs is performed, with the synthesized facial images being used, together with the original images, as DNN training and/or test data respectively; this also verifies the improved performance of our approach.
An ablation study is also presented, illustrating the effect of data granularity and subjects' age on the performance of the proposed method. Finally, conclusions and future work are presented in Section 6.

The proposed approach includes many novel contributions. To the best of our knowledge, it is the first time that the dimensional model of affect is taken into account when synthesizing face images. As verified in the experimental study, the generated images are of high quality and realistic. All other methods produce synthesized faces according to the six basic, or a few more, expressions. We further show that the proposed approach can accurately synthesize the six basic expressions.  

Moreover, it is the first time that a 4D face database is annotated in terms of valence and arousal and is then used for affect synthesis. The fact that this a temporal database ensures that successive video frames' annotations are adjacent in the VA Space. Consequently, we generate temporal affect sequences on a given neutral face by using annotations that are adjacent in the VA Space. Results are presented in the qualitative experimental study that illustrate this novel capability.

It should be also mentioned that the proposed approach works well, when presented with a neutral face image, obtained either in a controlled environment, or in-the-wild, e.g., irrespective of the head pose of the person appearing in the image. 

An extensive experimental study is provided, over most significant databases with affect, showing that the developed DNNs based on the proposed facial affect synthesis approach outperform the existing state-of-the-art, as well the same DNNs based on facial affect synthesis produced by GAN architectures.

\section{Related Work}\label{related_work}

Facial expression transfer is a research field for mapping and generating desired images of specified subject and facial expression. Many methods achieved significant results for high-resolution images and are applied to a wide range of applications, such as facial animation, facial editing, and facial expression recognition.

There are mainly two categories of methods for facial expression transfer from a single image: traditional graphic-based methods and emerging generative methods. In the first case, some methods directly warp the input face to create the targeted expression, by either 2D warps \cite{fried2016perspective,garrido2014automatic}, or 3D warps \cite{blanz2003reanimating,cao2014displaced,liu2008facial}. Other methods construct parametric global models. In \cite{mohammed2009visio}, a probabilistic model is learned, in which existing and generated images obey structural constraints. \cite{averbuch2017bringing} added fine‐scale dynamic details, such as wrinkles and inner mouth, that are associated with facial expressions. Although these methods have achieved some positive results in high-resolution and one-to-many image synthesis, they are still limited due to their sophisticated design and expensive computation.

In \cite{thies2016face2face}, the authors developed a real-time face-to-face expression transfer system, with an extra blending step for mouth. This 2D-to-3D approach shows promising results, but due to the nature of its formulation, it is unable to retrieve fine-details, and its applicability is limited to expressions
lying in a linear shape subspace with known rank.
The authors extended this system to human portrait video transfer~\cite{thies2018headon}. They captured facial expressions, eye gaze, rigid head pose, and motions of the upper body of a source actor and transferred them to a target actor in real time.

The second category of methods is based on data-driven generative models. At the beginning, some  generative models, such as deep belief nets (DBN)\cite{susskind2008generating} and higher-order Boltzmann machines \cite{reed2014learning}, had been applied to facial expression synthesis. However, these models faced problems such as blurry generated images, incapability of fine control of facial expression and low-resolution outputs.

With the recent development of Generative Adversarial
Networks (GANs) \cite{goodfellow2014generative}, these networks have been applied to facial expression transfer; due to the fact that the generated images are of high-quality, these provided positive results. A generative model is trained according to a dataset, including all information about identity, expression, viewing angle, etc, while performing facial expression transfer. Generative modeling methods reduce the complicated design of the connection between facial textures and emotional states and encode intuitionistic facial features into parameters of data distribution. However, the main drawback of GANs is the training instability and the trade-off between visual quality and image diversity.

Since the original GAN could not generate facial images with a specific facial expression referring to a specific person, some methods conditioned on expression categories have been proposed. Conditional GANs (cGANs) \cite{mirza2014conditional} (and conditional variational autoencoders (cVAEs) \cite{sohn2015learning}) can generate samples conditioned on attribute information, when this is available. Those networks require large training databases so that identity information can be properly disambiguated. Otherwise, when presented with an unseen face, the networks tend to generate faces which look like the “closest” subject in the training datasets. During training, those networks require the knowledge of the attribute labels; it is not clear how to adapt them to new attributes without retraining from scratch. Finally, these networks suffer from mode-collapse (e.g., the generator only outputs samples from a single mode, or with extremely low variety) and blurriness.

The conditional difference adversarial 
autoencoder (CDAAE) \cite{zhou2017photorealistic} aims at 
synthesizing specific expressions for unseen persons with a targeted emotion or facial action unit label. However, such GAN-based methods are still limited to discrete facial expression synthesis, i.e., they cannot generate a face sequence showing a smooth transition from an emotion to another. \cite{ding2018exprgan} proposed an Expression Generative Adversarial Network (ExprGAN) in which the expression intensity could be controlled in a continuous manner from weak to strong. The identity and expression representation learning were disentangled and there was no rigid requirement of paired samples for training. The authors developed a three-stage incremental learning algorithm to train the model on small datasets.

In \cite{pham2018generative}, the authors proposed a weakly supervised adversarial learning framework for automatic facial expression synthesis based on continuous action unit coefficients. In \cite{pumarola2018ganimation}, the Ganimation was proposed that additionally controlled the generated expression by AU labels, and allowed a continuous expression transformation. In addition, the authors introduced an attention-based generator to promote the robustness of their model for distracting backgrounds and illuminations.

Recently, \cite{song2018geometry} utilized landmarks and proposed the geometry-guided GAN (G2GAN) to generate smooth image sequences of facial expressions. G2GAN uses geometry information based on dual adversarial networks to express face changes and synthesizes facial images. Through manipulating  landmarks, smoothly changed images can also be generated. However, this method demands a neutral face of the targeted person as the intermediate of facial expression transfer. Although the expression removal network could generate a neutral expression of a specific person, this procedure brings additional artifacts and degrades the performance of expression transition.

\cite{qiao2018geometry} used geometry (facial landmarks) to control the expression synthesis with a facial geometry embedding network and proposed a Geometry-Contrastive Generative Adversarial Network (GC-GAN) to transfer continuous emotions across different subjects, even if there existed big difference in shapes. \cite{wu2018reenactgan} proposed a boundary latent space and boundary transformer. They mapped the source face into the boundary latent space, and transformed the source face’s boundary to the target’s boundary, which was the medium to capture facial geometric variances during expression transfer.

In~\cite{ma2019real}, an unpaired learning framework was developed to learn the mapping between any two facial expressions in the facial blendshape space. This framework automatically transforms the source expression in an input video clip to a specified target expression. This work lacks the capability to generate personalized expressions; individual-specific expression characteristics, such as wrinkles and creases, are ignored. Also, the transitions between different expressions are not taken into consideration. Finally, this work is limited in the sense that it cannot produce highly exaggerated expressions.

Both the graphic-based methods and the genererative methods of facial expression transfer have been used to create synthetic data that are used as auxiliary data in network training, augmenting the training dataset. A synthetic data generation system with a 3D convolutional neural network (CNN) was created in \cite{abbasnejad2017using} to confidentially create faces with different levels of saturation in expression. 
\cite{antoniou2017data} proposed the Data Augmentation
Generative Adversarial Network (DAGAN) which is based on  cGAN and tested its effectiveness on vanilla classifiers and one shot learning. DAGAN is a basic framework for data augmentation based on cGAN.

\cite{zhu2018emotion} presented another basic framework for face data augmentation based on CycleGAN\cite{zhu2017unpaired}. Similar to cGAN, CycleGAN is also an general-purpose solution for image-to-image translation, but it learns a dual mapping between two domains simultaneously with no need for paired training examples, because it combines a cycle consistency loss with adversarial loss. The authors used this framework to generate auxiliary data for imbalanced datasets, where the data class with fewer samples was selected as transfer target and the data class with more samples was the reference. %+ isws prosthese paper: DATA AUGMENTATION FOR LOW RESOURCE SENTIMENT ANALYSIS USING GENERATIVE ADVERSARIAL NETWORKS

\section{Materials \& Methods}\label{materials}

In the following, we first describe the 4DFAB database, its annotation in terms of valence-arousal and the selection of expressive categorical sequences from it. The annotated 4DFAB database has been used for constructing the 3D facial expression gallery that is the basis of our affect synthesis pipeline described in the next Section. Then we describe the methods we have used: a) for registering and correlating all components of the 3D gallery into a universal coordinate frame; b) for constructing the 3D Morphable Model used in this work.

\subsection{The 4DFAB Database}
\label{subsec:4dfab}
The 4DFAB database~\cite{cheng4dfab} is the first large scale 4D face database designed for biometric applications and facial expression analysis. It consists of 180 subjects (60 females, 120 males) aging from 5 to 75 years. 4DFAB was collected over a period of 5 years under four different sessions, with over 1,800,000 3D faces. The database was designed to capture articulated facial actions and spontaneous facial behaviors, the latter being elicited by watching emotional video clips. In each of the four sessions, different video clips were shown that stimulated different spontaneous behaviors. In this paper, we use all 1,580 spontaneous expression sequences (video clips) for dimensional emotion analysis and synthesis. The frame rate of 4DFAB database is 60 FPS and the average clip length for spontaneous expression sequences is 380 frames. Consequently the 1,580 expression sequences correspond to 600,000 frames, which we annotated in terms of valence and arousal (details follow in the next subsection). These sequences cover a wide range of expressions as shown in Figs. \ref{2d_wheel} and \ref{hist_4dfab}.

Moreover, to be able to develop the categorical emotion synthesis model, we used the 2,000 expressive 3D meshes per basic expression (12,000 meshes in total) that were provided along with 4DFAB. Those 3D meshes corresponded to (annotated) apex frames of posed expression sequences in
4DFAB. Such examples are shown in Fig. 1.   %%%%%%%%%%%%%%% to EXW VALEI MANUALLY EDW, ALLAXE TO

\begin{figure}[h]
\label{apex}
\centering
\scalebox{0.75}{
\begin{tabular}{ |c|c| }
 \hline
AN & \adjincludegraphics[height=1.2cm]{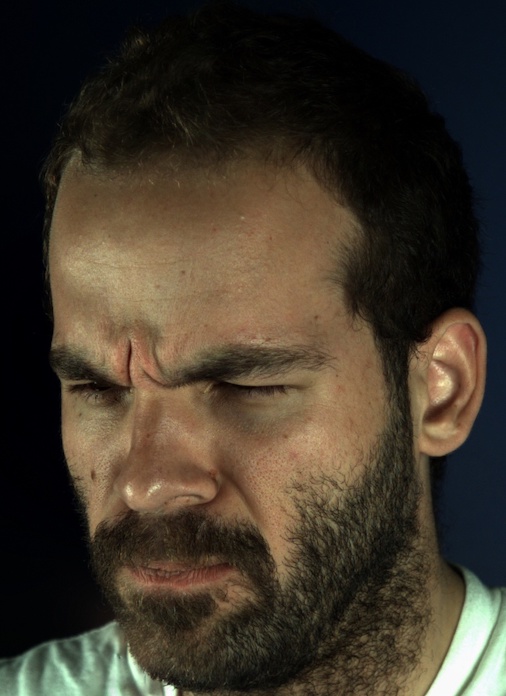}  \adjincludegraphics[height=1.2cm]{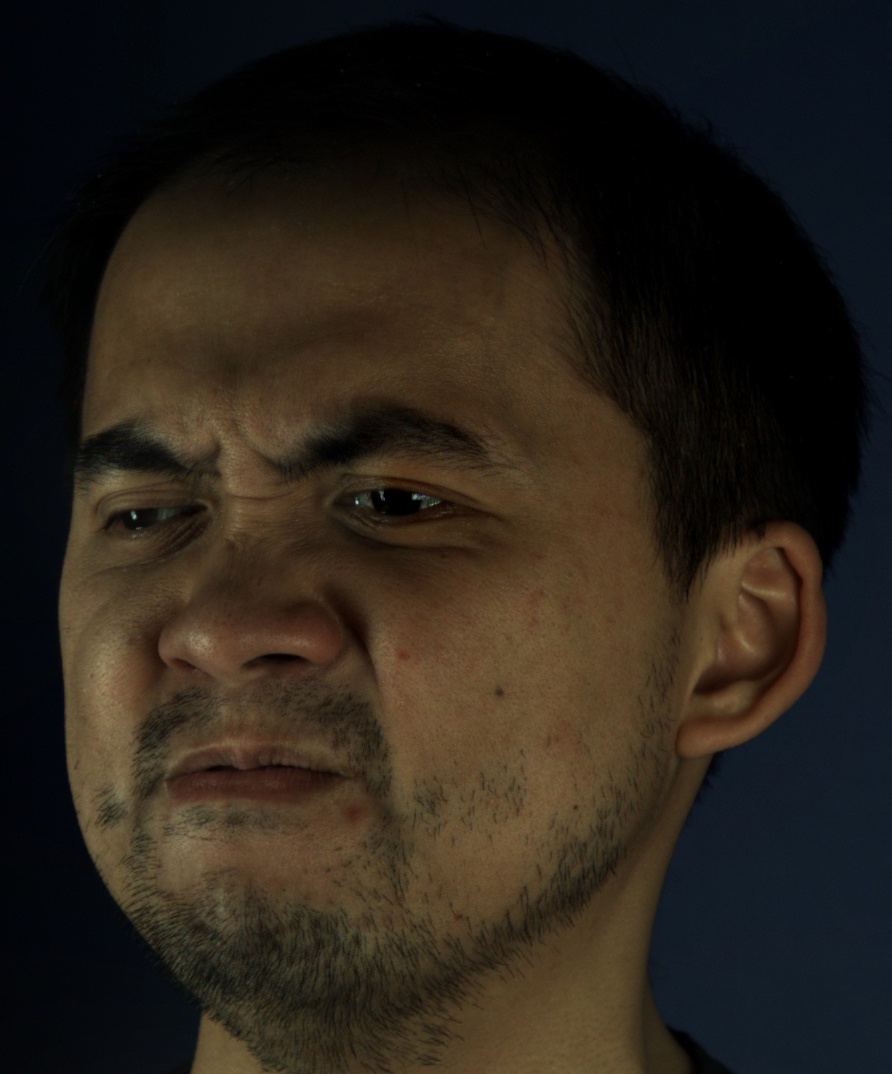}  \adjincludegraphics[height=1.2cm]{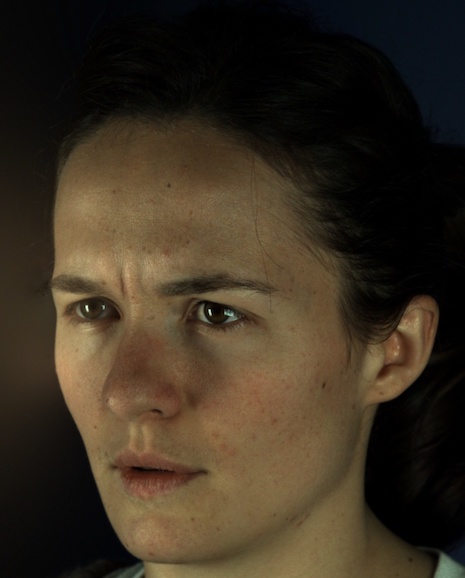}  \adjincludegraphics[height=1.2cm]{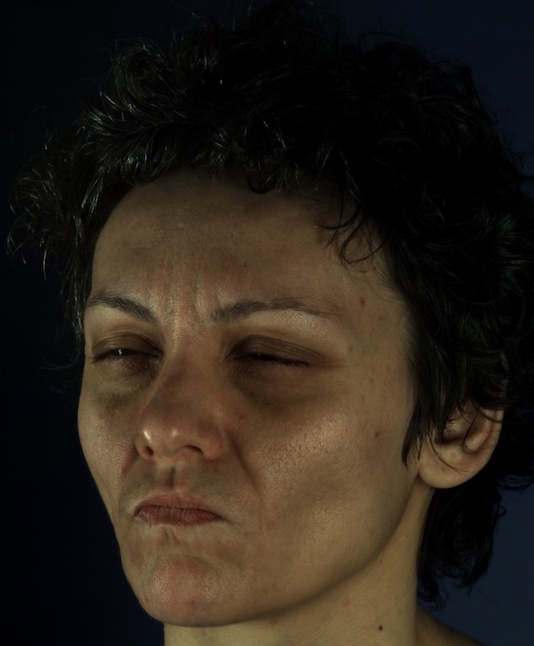}  \adjincludegraphics[height=1.2cm]{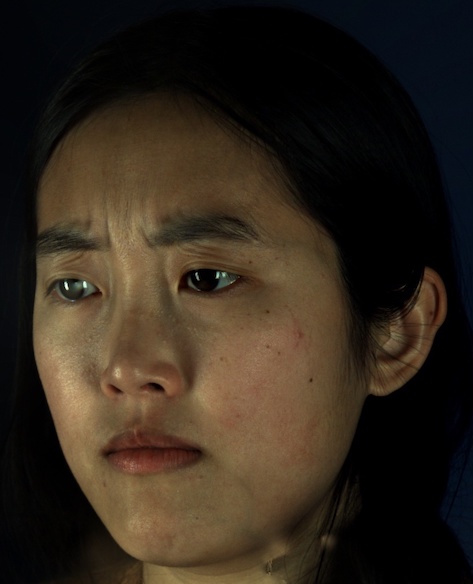}  \adjincludegraphics[height=1.2cm]{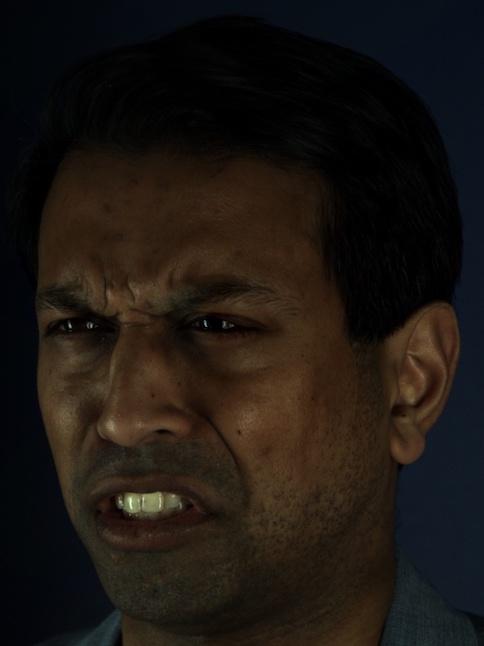}  \adjincludegraphics[height=1.2cm]{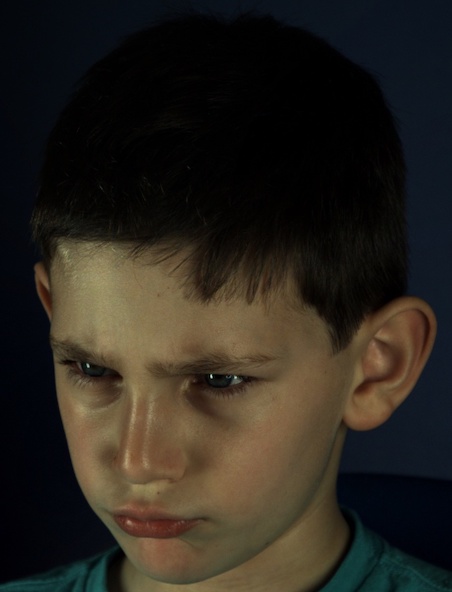} \\
\hhline{|-|-|} 
DI & \adjincludegraphics[height=1.2cm]{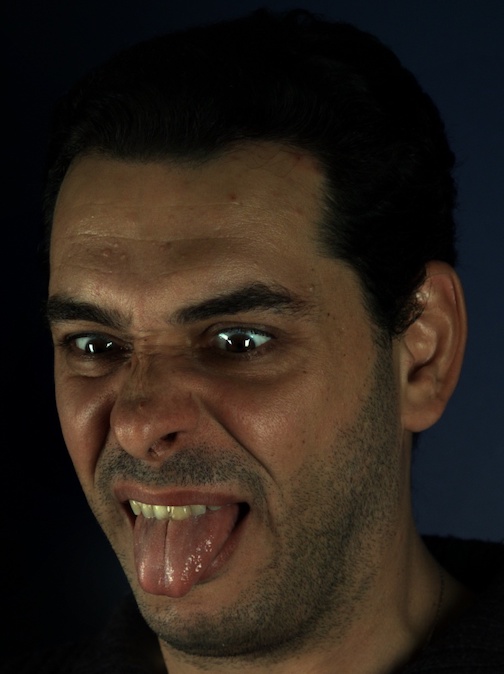}  \adjincludegraphics[height=1.2cm]{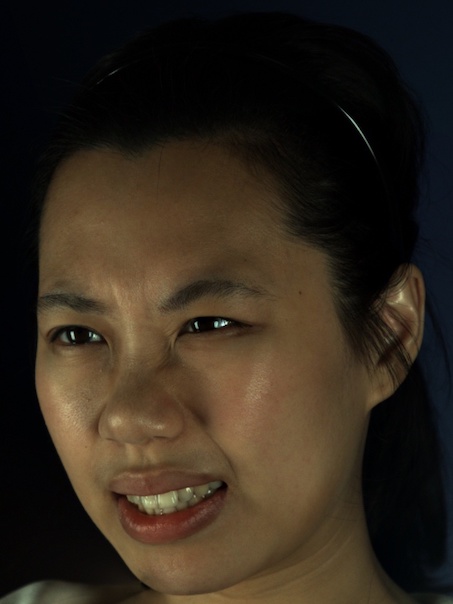}  \adjincludegraphics[height=1.2cm]{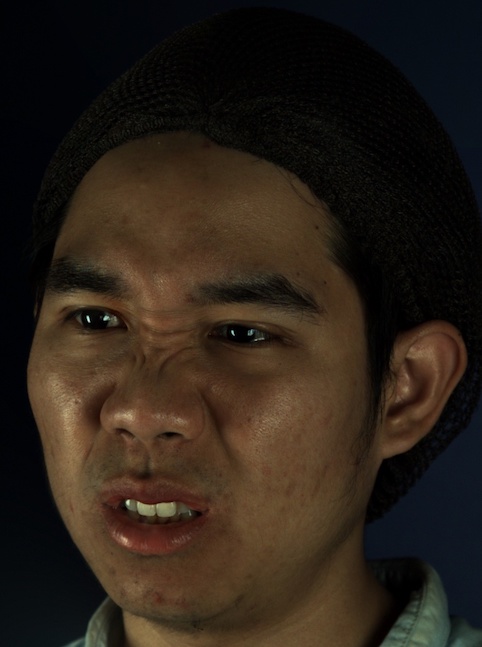}  \adjincludegraphics[height=1.2cm]{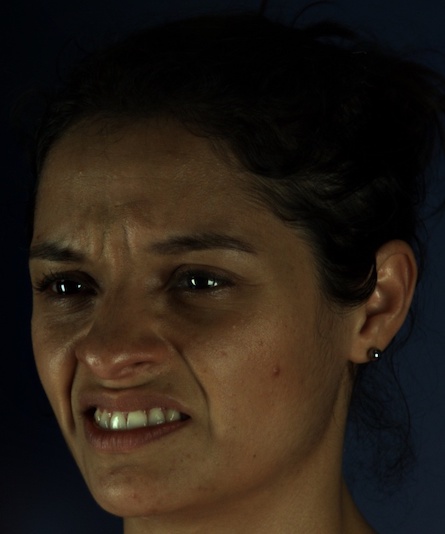}  \adjincludegraphics[height=1.2cm]{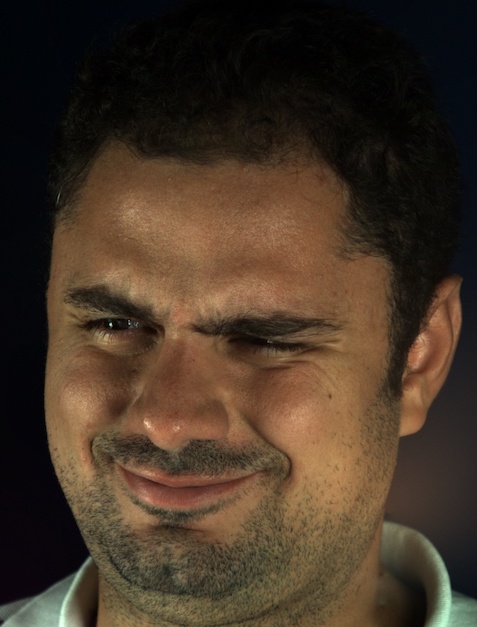}  \adjincludegraphics[height=1.2cm]{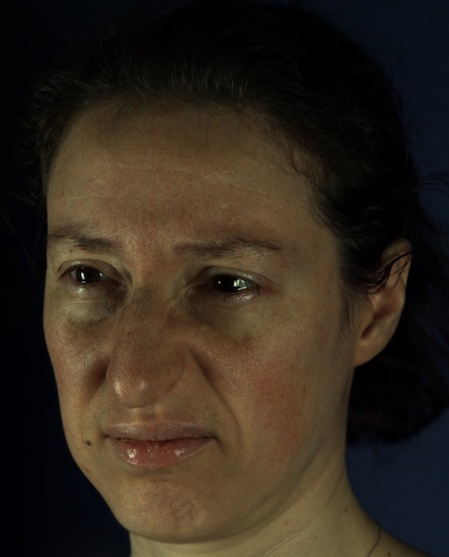}  \adjincludegraphics[height=1.2cm]{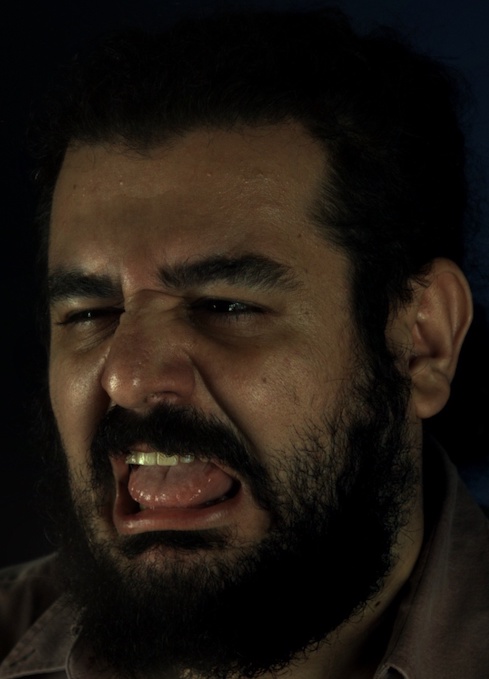} \\
\hhline{|-|-|} 
FE & \adjincludegraphics[height=1.2cm]{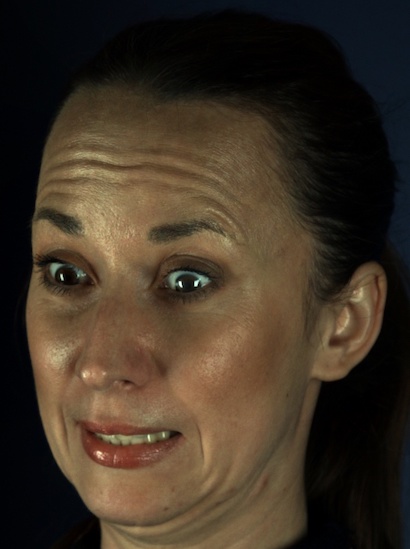}  \adjincludegraphics[height=1.2cm]{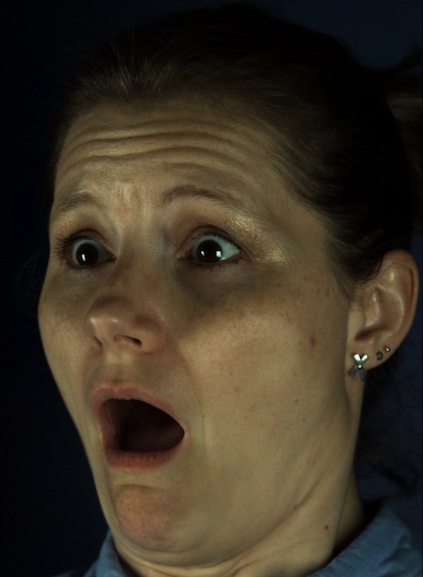}  \adjincludegraphics[height=1.2cm]{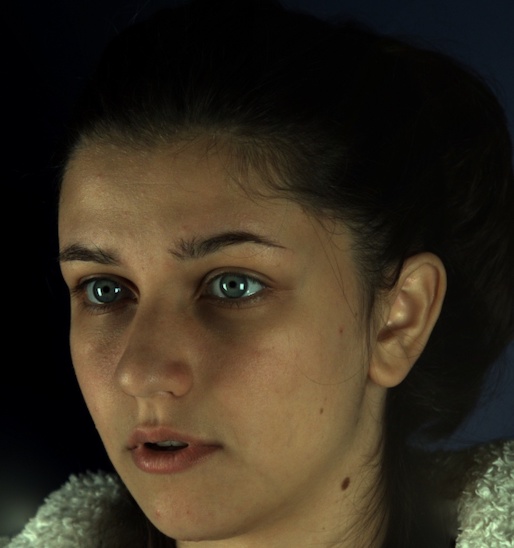}  \adjincludegraphics[height=1.2cm]{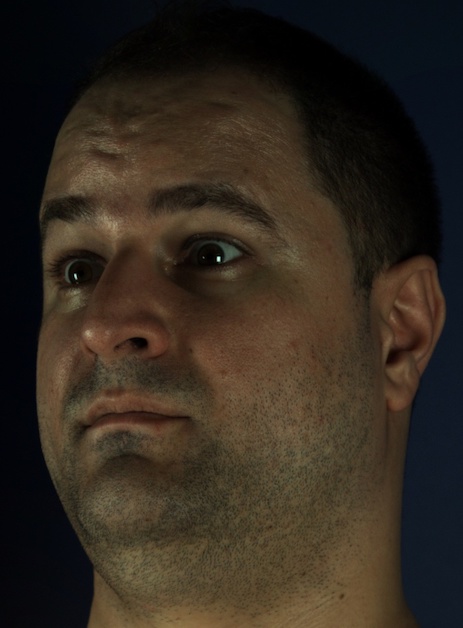}  \adjincludegraphics[height=1.2cm]{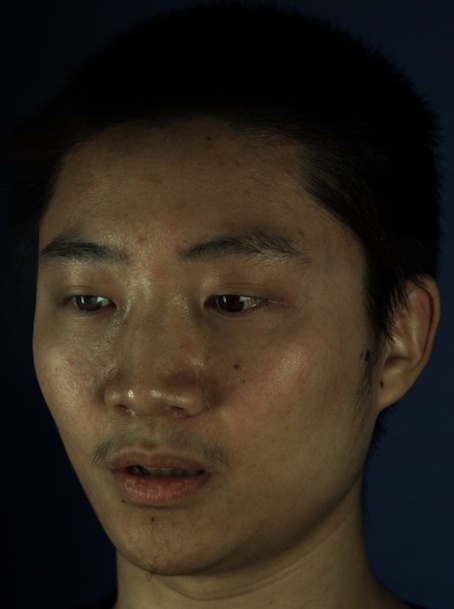}  \adjincludegraphics[height=1.2cm]{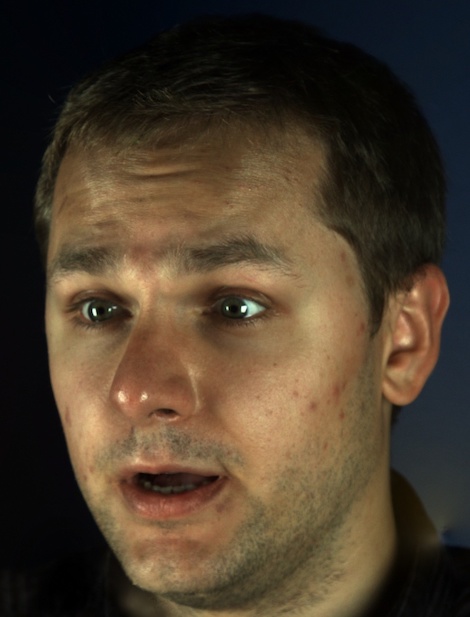}  \adjincludegraphics[height=1.2cm]{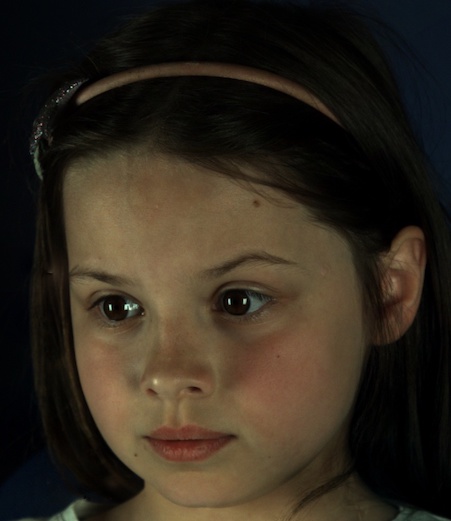} \\
\hhline{|-|-|} 
J & \adjincludegraphics[height=1.2cm]{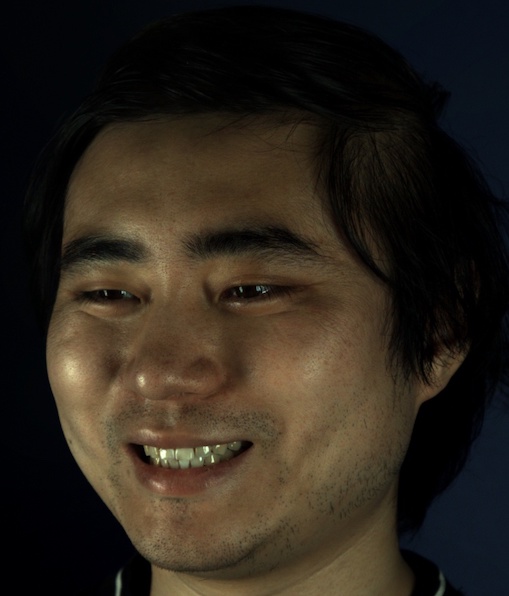}  \adjincludegraphics[height=1.2cm]{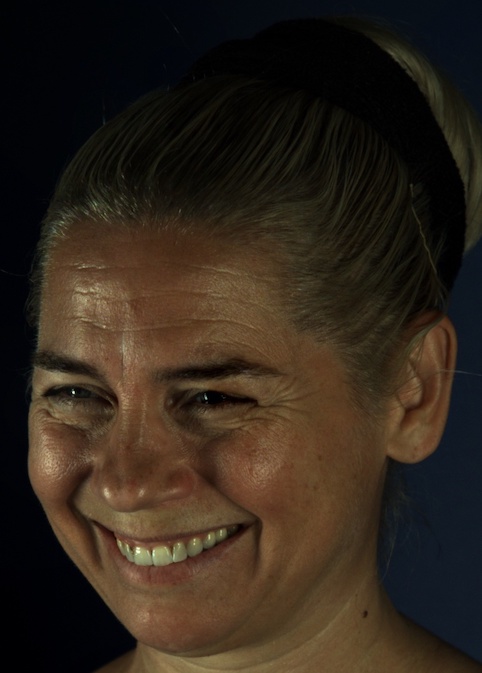}  \adjincludegraphics[height=1.2cm]{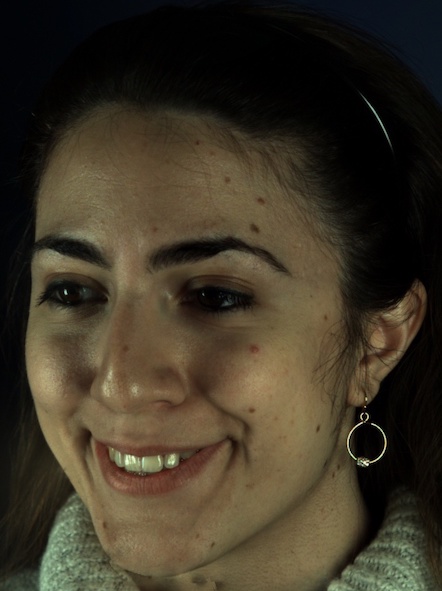}  \adjincludegraphics[height=1.2cm]{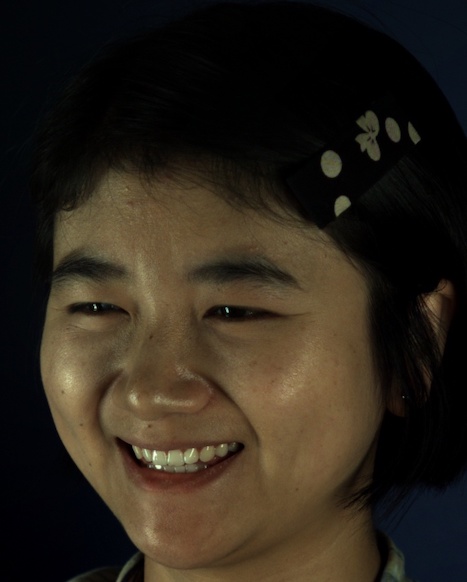}  \adjincludegraphics[height=1.2cm]{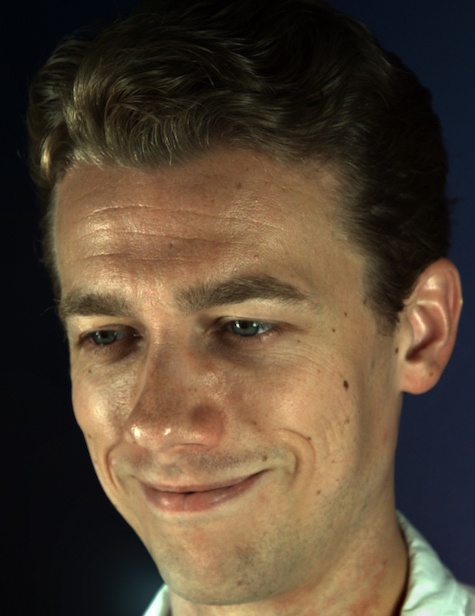}  \adjincludegraphics[height=1.2cm]{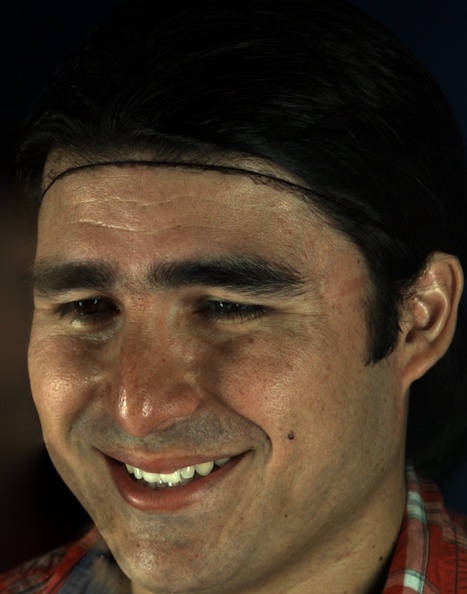}  \adjincludegraphics[height=1.2cm]{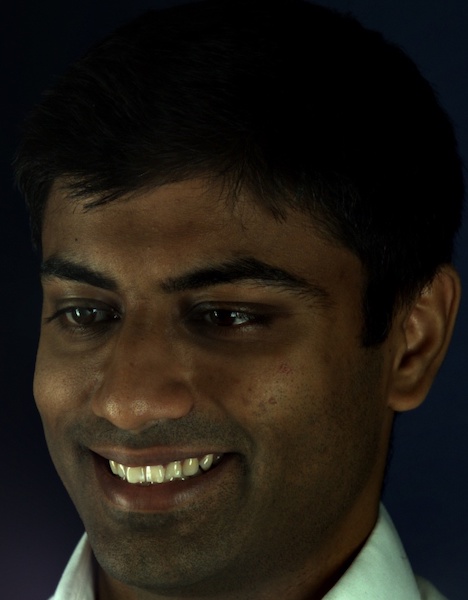} \\
\hhline{|-|-|} 
SA & \adjincludegraphics[height=1.2cm]{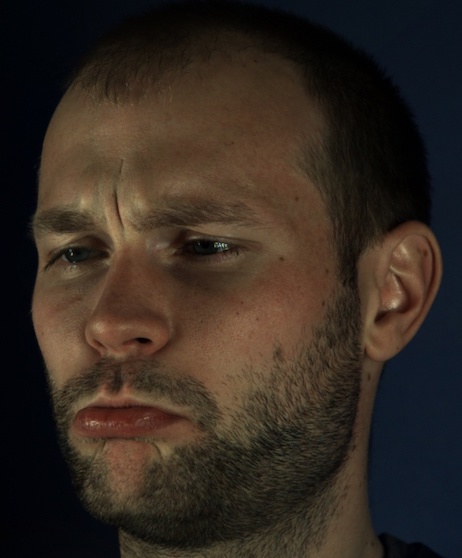}  \adjincludegraphics[height=1.2cm]{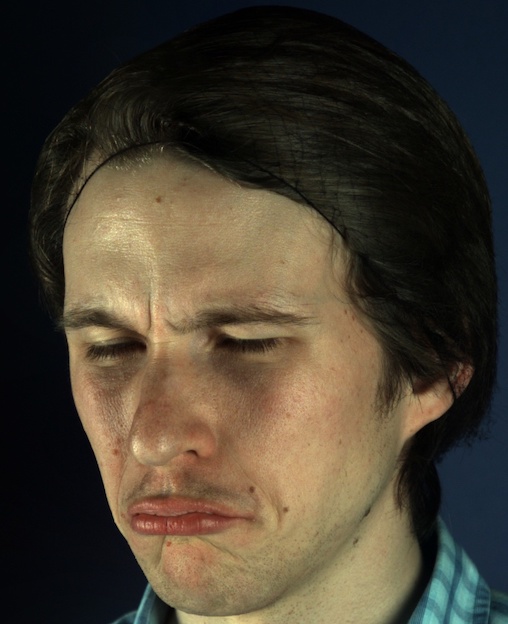}  \adjincludegraphics[height=1.2cm]{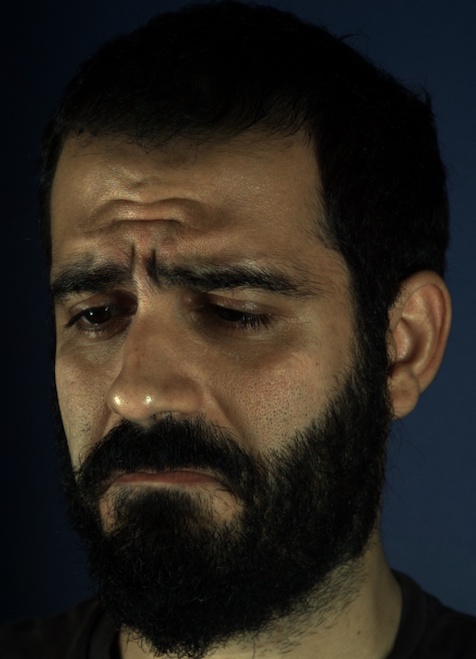}  \adjincludegraphics[height=1.2cm]{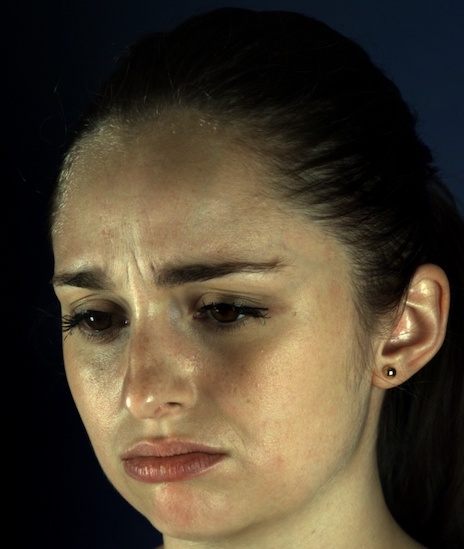}  \adjincludegraphics[height=1.2cm]{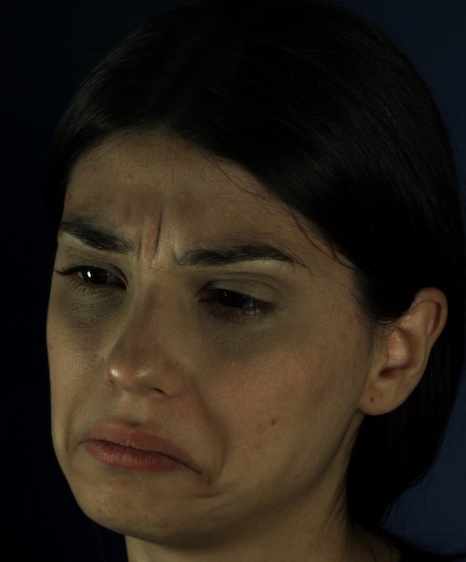}  \adjincludegraphics[height=1.2cm]{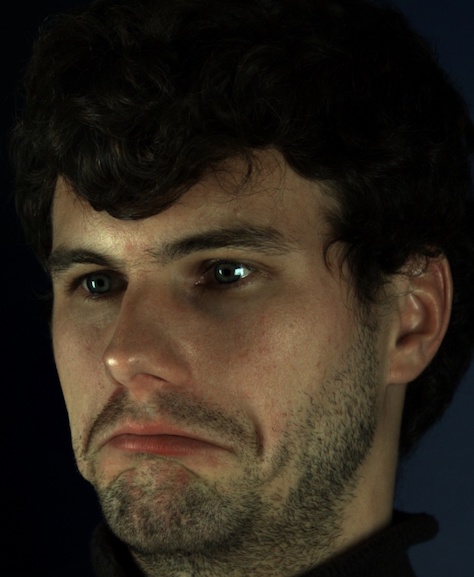}  \adjincludegraphics[height=1.2cm]{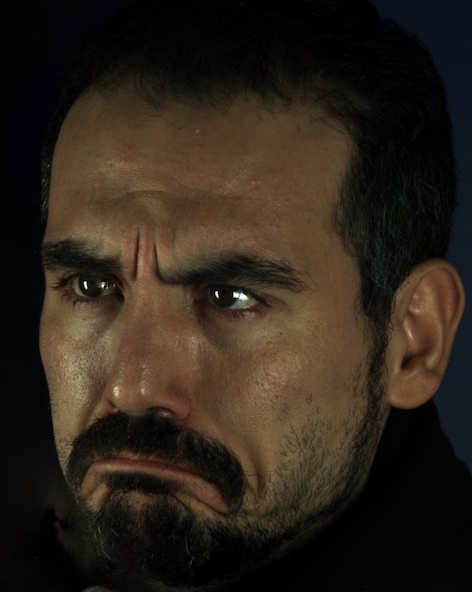} \\
\hhline{|-|-|} 
SU & \adjincludegraphics[height=1.2cm]{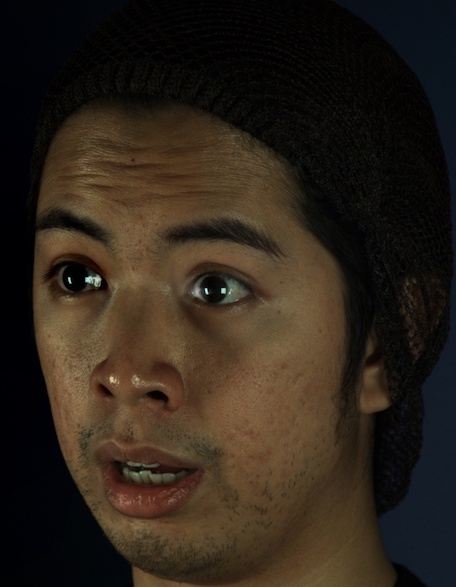}  \adjincludegraphics[height=1.2cm]{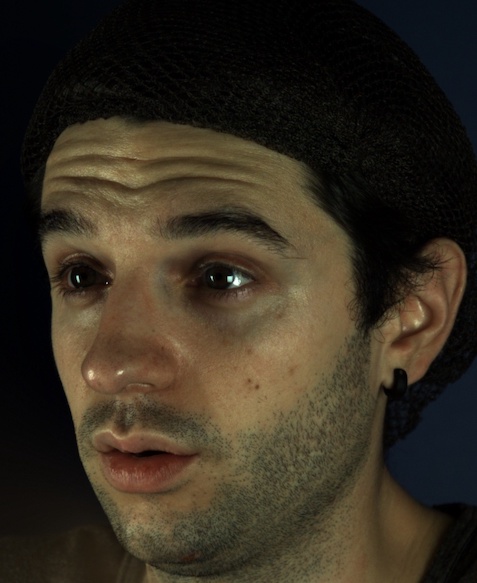}  \adjincludegraphics[height=1.2cm]{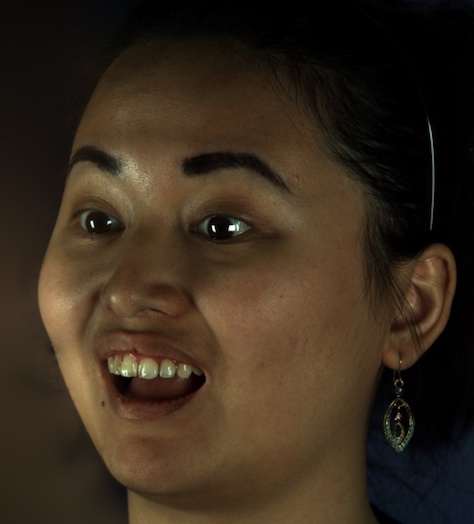}  \adjincludegraphics[height=1.2cm]{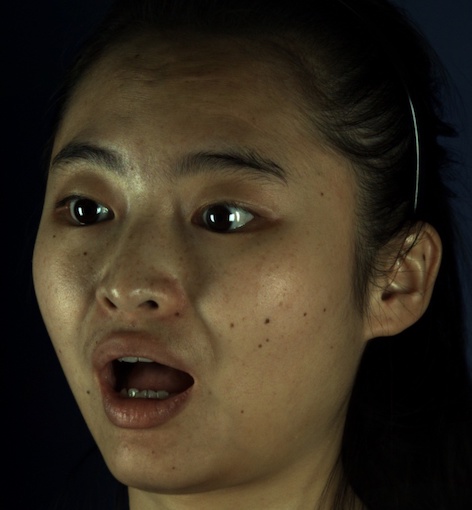}  \adjincludegraphics[height=1.2cm]{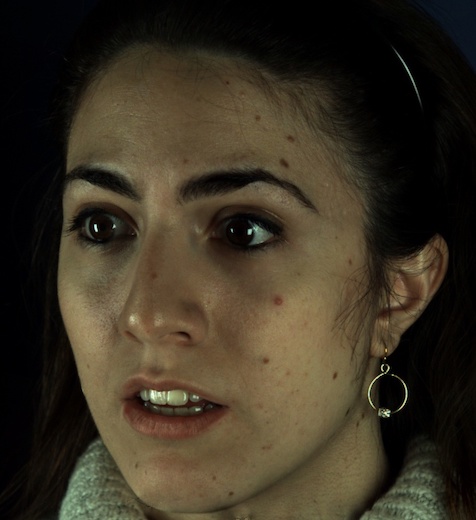}  \adjincludegraphics[height=1.2cm]{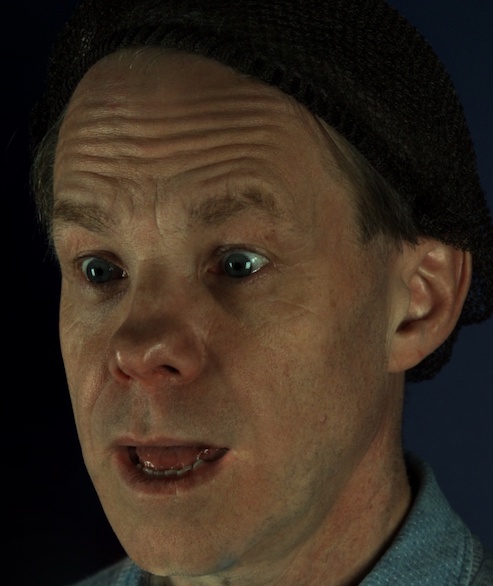}  \adjincludegraphics[height=1.2cm]{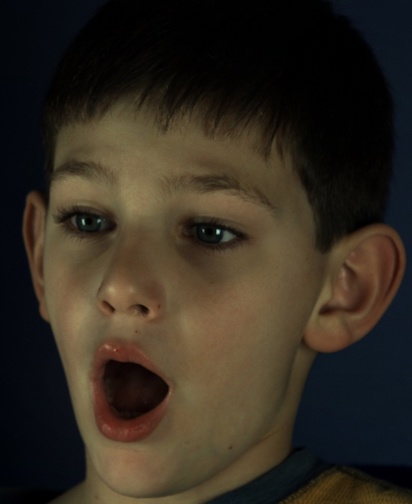} \\
\hline
\end{tabular}
}
\caption{Examples from the 4DFAB of apex frames with posed expressions for the six basic expressions: Anger (AN), Disgust (DI), Fear (FE), Joy (J), Sadness (SA), Surprise (SU)}
\end{figure}

\subsection{4DFAB Dimensional Annotation}

Targeting to develop the novel dimensional expression synthesis method, all 1,580 dynamic 3D sequences (i.e., over 600,000 frames) of 4DFAB have been annotated in terms of valence and arousal emotion dimensions. In total, three experts were chosen to perform the annotation task. Each expert performed a time-continuous annotation for both affective dimensions. The application-tool described in \cite{zafeiriou2017aff}, was used in the annotation process.

Each expert logged into the application-annotation tool using an identifier (e.g. his/her name) and selected an appropriate joystick; then the application showed a scrolling list of all videos and the expert selected a video to annotate; then a screen appeared that showed the selected video and a slider of valence or arousal values ranging in $[-1,1]$ ; the expert annotated the video by moving the joystick either up or down; finally, a file was created with the annotations. 
The mean inter-annotation correlation per annotator was 0.66, 0.70, 0.68 for valence and 0.59, 0.62, 0.59 for arousal. The average of those mean inter-annotation correlations was 0.68 for valence and 0.60 for arousal. Those values are high, indicating a very good agreement between annotators. As a consequence, the final label values were chosen to be the mean of those three annotations. 

\begin{figure}[h!]
\centering
\adjincludegraphics[height=6cm,width=8cm,trim={{.09\totalheight} {.06\totalheight} {.07\totalheight} {.08\totalheight}}]{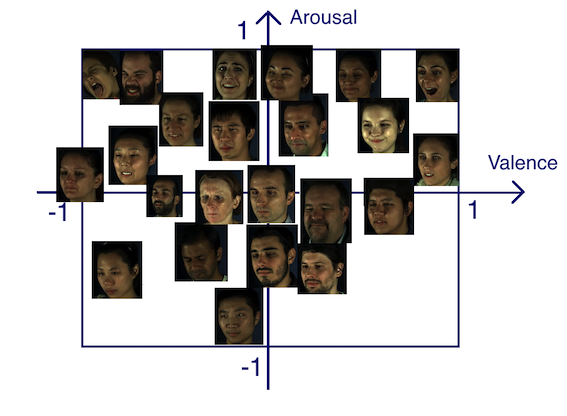}  
\caption{The 2D Valence-Arousal Space and some representative frames of 4DFAB}
\label{2d_wheel}
\end{figure}

\begin{figure}[h!]
\centering
\adjincludegraphics[height=6.5cm]{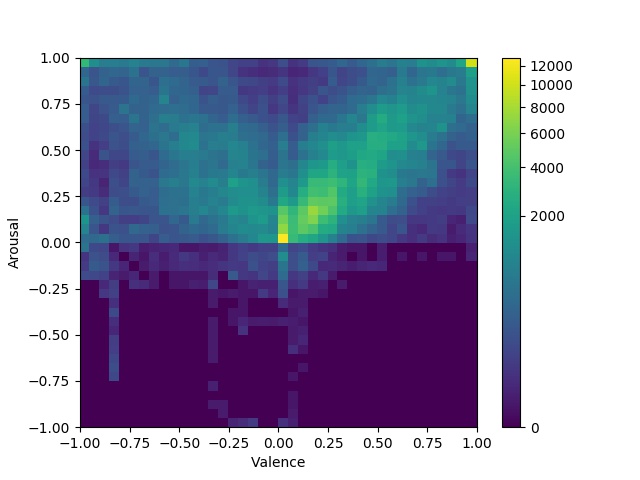}  
\caption{The 2D histogram of annotations of 4DFAB}
\label{hist_4dfab}
\end{figure}

Examples of frames from the 4DFAB along with their annotations, are shown in Fig.~\ref{2d_wheel}. Fig. \ref{hist_4dfab} shows the 2D histogram of annotations of 4DFAB. In the rest of the paper, we refer to the 4DFAB database either as: i) the 600,000 frames with their corresponding 3D meshes, which have been annotated with 2D valence and arousal (VA) emotion values or ii) the 12,000 apex frames of posed expressions with their corresponding 3D meshes, which have categorical annotation.

\subsection{Mesh Pre-Processing: Establishing Dense Correspondence} \label{dense_corr}

Each 3D mesh is first re-parameterized into
a consistent form where the number of vertices, the triangulation and the anatomical meaning of each vertex are made
consistent across all meshes. For example, if the vertex with
index i in one mesh corresponds to the nose tip, it is required
that the vertex with the same index in every mesh corresponds
to the nose tip too. Meshes satisfying the above properties are
said to be in dense correspondence with one another. So, correlating all these meshes with a universal coordinate frame (viz. a 3D face template) is a step to follow so as to establish dense correspondence. 

In order to do so, we need to define a 2D \emph{UV} space
for each mesh, which in fact is a contiguous flattened atlas that embeds the 3D facial surface. Such a UV space is associated with its corresponding 3D surface through a bijective mapping; thus, establishing dense correspondence between two UV images implicitly establishes a 3D-to-3D correspondence for the mapped mesh. UV mapping is the 3D modelling process of projecting a 2D image to a 3D model's surface for texture mapping. The letters \emph{U} and \emph{V} denote the axes of the 2D texture, since \emph{X, Y} and \emph{Z} are already taken to denote the axes of the 3D object in model space. 

We employ an optimal cylindrical projection method \cite{booth2014optimal} to synthetically create a UV space for each mesh. A UV map (which is an image I), with each pixel encoding both spatial information (X, Y, Z) and texture information (R, G, B), is produced, on which we perform non-rigid alignment. Non-rigid alignment is performed through the UV-TPS method that utilises key landmarks fitting and Thin Plate Spline (TPS) warping \cite{cosker2011facs}. Following \cite{cheng4dfab}, we perform several modifications to \cite{cosker2011facs}, to suit our data. Firstly, we build session-and-person-specific Active Appearance Models (AAMs)\cite{alabort2017unified} to automatically track feature points in the UV sequences. This means that 4 different AAMs are built and used separately for one subject. Main reasons behind this are: i) textures of different sessions differ due to several facts (i.e. aging, beards, make-ups, experiment lighting condition), ii) person-specific model is proven more accurate and robust in specific domains \cite{chew2012pursuit}. 

In total, 435 neutral meshes and 1047 expression meshes (1 neutral and 2-3 expressive meshes per person and session) in 4DFAB were selected; these contained annotations with 79 3D landmarks.  They
were unwrapped and rasterised to UV space, then
grouped for building the corresponding AAMs. Each UV map was flipped to increase fitting robustness. Once all the UV sequences were tracked with 79 landmarks, they were then warped to the corresponding reference frame using TPS, thus achieving the 3D
dense correspondence. For each subject and session, one specific reference coordinate frame from his/her
neutral UV map was built. From each warped frame, we could uniformly sample the texture and 3D coordinates. Eventually, a set of non-rigidly corresponded 3D meshes under the same topology and density were obtained.

Given that meshes have been aligned to their designated reference frame, the last step was to establish dense 3D-to-3D correspondences between those reference frames and a 3D template face. This is a 3D mesh registration problem,  solved by Non-rigid ICP\cite{amberg2007optimal}. We employed it to register the neutral reference meshes to a common template, the Large Scale Facial Model (LSFM)~\cite{booth2018large}. We brought all 600,000 3D meshes into full correspondence with the mean face of LSFM. As a result, we created a new set of 600,000 3D faces that share identical mesh topology, while maintaining their original facial expressions. In the following, this set constitutes the 3D facial expression gallery which we use for facial affect synthesis.

\subsection{Constructing a 3D Morphable Model}

\subsubsection{General Pipeline}

A common 3DMM consists of three parametric models: the shape, the camera and the texture models.

To build the shape model, the training 3D meshes should be put in dense correspondence (similarly to the previous Mesh Pre-Processing subsection). Next, Generalized Procrustes Analysis is performed to remove any similarity effects, leaving only shape information. Finally, Principal Component Analysis (PCA) is applied to these meshes, which generates a 3D deformable model as a linear basis of shapes. This model allows for the generation of novel shape instances. The model can be expressed as:
\begin{align}
    \mathcal{S}(\mathbf{p}) =& \ \mathbf{\bar{s}} + \mathbf{U}_{s}\mathbf{p}
\label{eq_shape_model}
\end{align}
where $\mathbf{\bar{s}} \in \mathbb{R}^{3N}$ is the mean component of 3D shape (in our case it is the mean of shape models from the LSFM model described in the next subsection) with $N$ denoting the number of vertices in the shape model; $\mathbf{U}_{s} \in \mathbb{R}^{3N \times n_s}$ is the shape eigenbase (in our case it is the identity subspace of LSFM) with $n_s << 3N$ being the number of principal components ($n_s$ is chosen to explain a percentage of the training set variance; generally, this percentage is 99.5\%); and $\mathbf{p} \in \mathbb{R}^{n_s}$ is a vector of parameters which allows for the generation of novel shape instances.

The purpose of camera model is to project the object-centered Cartesian coordinates of a 3D mesh instance into 2D Cartesian coordinates in an image plane. At first, given that the camera is static, the 3D mesh is rotated and translated using a linear view transformation, which results in 3D rotation and translation components. Then, a nonlinear perspective transformation is applied. Note that quaternions \cite{kuipers1999quaternions,wheeler1995iterative} are used to parametrise the 3D rotation, which ensures computational efficiency, robustness and simpler differentiation. In this manner we construct the camera parameters (i.e., 3D translation components, quaternions and parameter of linear perspective transformation). The camera model of the 3DMM applies the above transformations on the 3D shape instances generated by the shape model. Finally, the camera model can be written as:
\begin{equation}
    \mathcal{W}(\mathbf{p}, \mathbf{c}) = \mathcal{P}(\mathcal{S}(\mathbf{p}), \mathbf{c}),
\label{eq_camera_model}
\end{equation}
where $\mathcal{S}(\mathbf{p})$ is a 3D face instance; $\mathbf{c} \in \mathbb{R}^{n_c}$ are the camera parameters (for rotation, translation and focal length; $n_c$ is 7); and $\mathcal{P} : \mathbb{R}^{3N} \rightarrow \mathbb{R}^{2N}$ is the perspective camera projection.

For the texture model, large facial “in-the-wild” data-bases annotated for sparse landmarks are needed. Let us assume that the meshes have corresponding camera and shape parameters. These images are passed through a dense feature extraction function that returns feature-based representations for each image. These are then sampled from the camera model at each vertex location so as to build a texture sample, which will be nonsensical for some regions mainly due to self occlusions present in the mesh projected in the image space. To complete the missing information of the texture samples, Robust PCA (RPCA) with missing values \cite{Shang2014RPCA} is applied. This produces complete feature-based textures that can be processed with PCA to create the statistical model of texture, which can be written as:
\begin{equation}
    \mathcal{T}(\boldsymbol{\lambda}) = \mathbf{\bar{t}} + \mathbf{U_t}\boldsymbol{\lambda},
\label{eq_texture_model}
\end{equation}
where $\mathbf{\overline{t}} \in \mathbb{R}^{3N}$ is the mean texture component (in our case it is the mean of texture model from LSFM); $\mathbf{U}_t \in \mathbb{R}^{3N \times n_t}$ and $\boldsymbol{\lambda} \in \mathbb{R}^{n_t}$ are the texture subspace (eigenbase) and texture parameters, respectively, with $n_t << 3N$ being the number of principal components. This model can be used to generate novel 3D feature-based texture instances.

\begin{figure*}[h!]
\centering
\adjincludegraphics[width=0.9\linewidth]{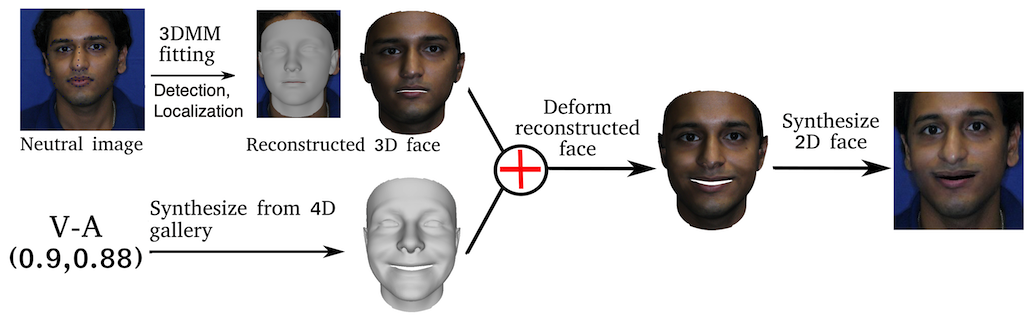}
\caption{The facial affect synthesis framework: the user inputs an arbitrary 2D neutral face and the affect to be synthesized (a pair of valence-arousal values in this case)}
\label{framework}
\end{figure*}

\subsubsection{The Large Scale Facial Model (LSFM)} 
We have adopted the LSFM model constructed using the MeIn3D dataset~\cite{booth2018large}. The construction pipeline of LSFM starts with a robust approach to 3D landmark localization resulting in generating 3D landmarks for the meshes. The 3D landmarks are then employed as soft constraints in Non-rigid ICP to place all meshes in correspondence with a template facial surface; the mean face of the Basel Face Model~\cite{paysan20093d} has been chosen. However, the large cohort of data could result in convergence failures. These are an unavoidable byproduct of the fact that both landmark localization and NICP are non-convex optimization problems sensitive to initialization. 

A refinement post-processing step weeds out problematic subjects automatically, guaranteeing that the LSFM models are only constructed from training data for which there exist a high confidence of successful processing. Finally, the LSFM models are derived by applying PCA on the corresponding training sets, after excluding the shape vectors that have been classified as outliers. In total, 9,663 subjects are used to build LSFM, which covers a wide variety of age (from 5 to over 80s), gender (48\% male, 52\% female), and ethnicity (82\% White, 9\% Asian, 5\% Mixed Heritage, 3\% Black and 1\% other).

\section{The Proposed Approach}\label{methodology}

In this Section, we present the fully automatic facial affect synthesis framework. The user needs to provide a neutral image and an affect, which can be a VA pair of values, a path in the 2D VA space, or one of the basic expression categories. Our approach: 1) performs face detection and landmark localization on the input neutral image, 2) fits  a 3D Morphable Model (3DMM) on the resulting image \cite{booth2017itw3dmm}, 3) deforms the reconstructed face and adds the input affect, and 4) blends the new face with the given affect into the original image. Here let us note that the total time needed for the first two steps is about 400ms; this has to be performed only once, if generating multiple images from the same input image. Specific details regarding the described steps of our approach follow. This procedure is shown in Fig. \ref{framework}.

\subsection{Face Detection \& Landmark Localization} 

The first step to edit an image is to locate landmark points that will be used for fitting the 3DMM. We first perform face detection with the face detection model from \cite{Zhang2016Signal} and then utilize \cite{Deng2018FG} to localize 68 2D facial landmark points which are aware of the 3D structure of the face, in the sense that points on occluded parts of the face (most commonly part of the jawline) are correctly localized.

\subsection{3DMM-Fitting: Cost Function \& Optimization} 

The goal of this step is to retrieve a reconstructed 3D face with the texture sampled from the original image. In order to do so, we first need a 3DMM; we select the LSFM. 

Fitting a 3DMM on face images is an inverse graphics approach to 3D reconstruction and consists of optimizing three parametric models of the 3DMM, the \textit{shape}, \textit{texture} and \textit{camera} models. The optimization aims at rendering a 2D image which is as close as as possible to the input one. 
In our pipeline we follow the 3DMM fitting approach of \cite{booth2017itw3dmm}. As is already noted, we employ the LSFM \cite{booth2018large}  
$\mathcal{S}(\mathbf{p})$ to model the identity deformation of faces. Moreover, we adopt the robust, feature-based texture model $\mathcal{T}(\boldsymbol{\lambda})$ of \cite{booth2017itw3dmm}, built from in-the-wild images. The employed camera model is a perspective transformation $\mathcal{W}(\mathbf{p}, \mathbf{c})$, which projects shape $\mathcal{S}(\mathbf{p})$ on the image plane.

Consequently, the objective function that we optimize can be formulated as:

\begin{align}
    \argmin_{\mathbf{p}, \boldsymbol{\lambda}, \mathbf{c}} & \ \|\mathbf{F}(\mathcal{W}(\mathbf{p}, \mathbf{c})) - \mathcal{T}(\boldsymbol{\lambda})\|^2 + c_l\|\mathcal{W}_l(\mathbf{p}, \mathbf{c}) - \mathbf{s}_l\|^2    \nonumber
    \\
    &+ \ c_s\|\mathbf{p}\|^2_{\Sigma_s^{-1}} + c_t\|\boldsymbol\lambda\|^2_{\Sigma_t^{-1}},
\label{eq_3DMM_cost}
\end{align}

\noindent where the first term denotes the pixel loss between the feature based image $\mathbf{F}$ sampled at the projected shape's locations and the model generated texture; the second term denotes a sparse landmark loss between the image 2D landmarks and the corresponding 2D projected 3D points, where the 2D shape, $\mathbf{s}_l$, is provided by~\cite{Deng2018FG}; the rest two terms are regularization terms which serve as counter over-fitting mechanism, where $\mathbf{\Sigma}_{s}$ and $\mathbf{\Sigma}_{t}$ are diagonal matrices with the main diagonal being eigenvalues of the shape and texture models respectively; $c_l$, $c_s$ and $c_t$ are weights used to regularize the importance of each term during optimization and were empirically set to $10^5$, $3\times 10^6$ and $1$, respectively, following \cite{booth2017itw3dmm}.
Note also, that the 2D landmarks term is useful as it drives the optimization to converge faster.
Problem of Eq. \ref{eq_3DMM_cost} is solved by the Project-Out variation of Gauss-Newton optimization as formulated in \cite{booth2017itw3dmm}.

From the optimized models, the optimal shape instance constitutes the neutral 3D representation of the input face. Moreover, by utilizing the optimal shape and camera models, we are able to sample the input image at the projected locations of the recovered mesh and extract a UV texture, that we later use for rendering.

\subsection{Deforming Face \& Adding Affect}

Given an affect and an arbitrary 2D image $\mathbf{I}$, we first fit the LSFM to this image using the aforementioned 3DMM fitting method. After that, we can retrieve a reconstructed 3D face $\mathbf{s}_{orig}$ with the texture sampled from the original image (texture sampling is simply extracting image pixel value for each projected 3D vertex in image plane). Let us assume that we have created an affect synthesis model $\mathbf{M}_{Aff}$ that takes the affect as input and generates a new expressive face (denoted as $\mathbf{s}_{gen}$), i.e., $\mathbf{s} = \mathbf{M}_{Aff} (affect)$ (specific details regarding the generation of the expressive face, can be found in subsection \ref{blendshapes}). Next, we calculate the facial deformation $\Delta\mathbf{s}$ by subtracting the synthesized face $\mathbf{s}_{gen}$ from the LSFM template $\Bar{\mathbf{s}}$, i.e., $\Delta\mathbf{s} = \mathbf{s}_{gen} - \Bar{\mathbf{s}}$, and impose this deformation on the reconstructed mesh, i.e., $\mathbf{s}_{new} = \mathbf{s}_{orig} + \Delta\mathbf{s}$. Therefore, we obtain a 3D face (dubbed $\mathbf{s}_{new}$) with facial affect.

\subsection{Synthesizing 2D Face} 

The final step in our pipeline is to render the new 3D face $\mathbf{s}_{new}$ back to the original 2D image. To do that we employ the mesh that we have deformed according to the given affect, the extracted UV texture and the optimal camera transformation of the 3DMM. For rendering, we pass the three model instances to a renderer and we use as background the background of the input image. Lastly, the rendered image is fused with the original image via poisson blending \cite{Perez2003Siggraph} to smooth the boundary between foreground face and image background so as to produce a natural and realistic result. In our experiments, we used both a CPU-based renderer \cite{menpo14} and a GPU-based renderer \cite{Genova_2018_CVPR}. The GPU-based renderer greatly decreases the rendering time, as it needs 20ms to render a single image, while the CPU-based renderer needs 400ms.

\subsection{Synthesizing Expressive Faces with Given Affect} \label{blendshapes}

\begin{figure}[h]
\centering
\adjincludegraphics[height=2cm]{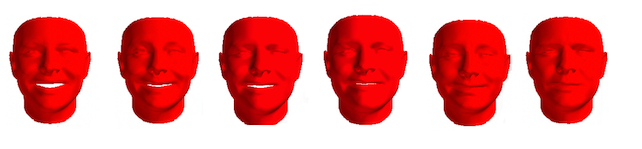}\\
\adjincludegraphics[height=2cm]{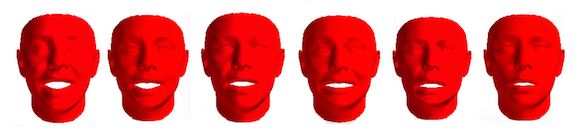}\\
\adjincludegraphics[height=2cm]{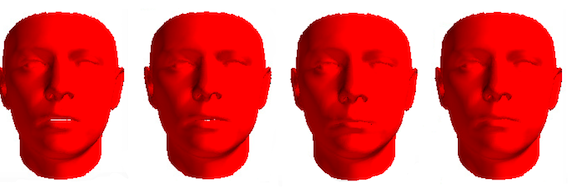}\\
\adjincludegraphics[height=2cm]{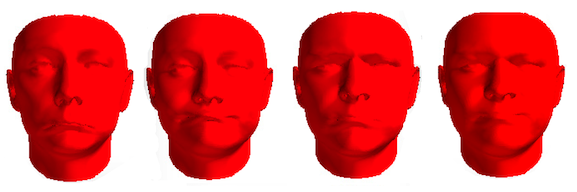}\\
\caption{Some mean faces of the 550 classes in the VA Space}
\label{blends}
\end{figure}

\begin{figure*}[h!]
\centering
\adjincludegraphics[width=0.99\linewidth]{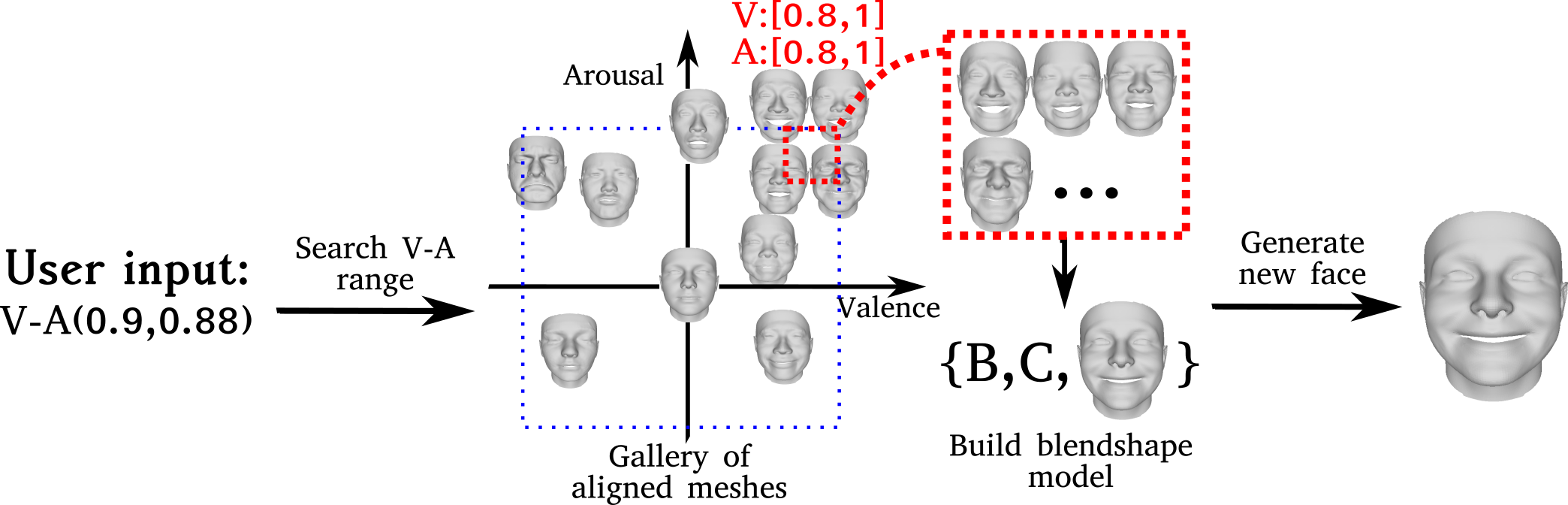}\\
\caption{Generation of new facial affect from the 4D face gallery; the user  provides a target VA pair}
\label{framework2}
\end{figure*}

\subsubsection{VA \& Basic Expression cases: Building Blendshape Models \& Computing Mean Faces} \label{va_mean_face}

Let us first describe the VA case. We have 600,000 3D meshes (established into dense correspondence) and their VA annotations. We want to appropriately discretize the VA Space into classes, so that each class contains a sufficient number of data. This is due to the fact that if classes contain only few examples, it is  more likely to include  identity information. However, the synthesized facial affect should only describe the expression associated with the VA pair of values, rather than information for the person's identity, gender, or age.
We have chosen to perform agglomerative clustering \cite{maimon2005data} on the VA values, using the euclidean distance as metric and the ward as linkage criterion (keeping the correspondence between VA values and 3D meshes). In this manner, we created 550 clusters, i.e.,  classes.  Then we built blendshape models and computed the mean face per class. Fig. \ref{blends} illustrates the mean faces of various classes. It should be mentioned that the majority of classes correspond to the first two quadrants of the VA Space, namely the regions of positive valence  (as can be seen in the 2D histogram of Fig. \ref{hist_4dfab}).

As far as the basic expression case is concerned, based on the derived  12,000 3D meshes, 2,000 for each of the six basic expressions, we built six blendshape models and six corresponding mean faces.

\subsubsection{User Selection: VA/Basic Expr \& Static/Temporal Synthesis} \label{choose}

The user first chooses the type of affect that our approach will generate. The affect could be either a point, or a path in the VA space, or one the six basic expression categories. If the user chooses the latter, then we retrieve the mean face of this category and add it on the 3D face reconstructed from the user's input neutral image. In this case, the only difference in Fig. \ref{framework} would be for the user to input a basic expression, the happy one, instead of a VA pair of values.
If the user chooses the former, then (s)he needs to additionally clarify if our approach should generate an image ('static synthesis') or a sequence of images ('temporal synthesis') with this affect. 

\bigskip

\noindent \textbf{Static synthesis} If the user selects 'static synthesis', then the user should input a specific VA pair of values. Then, we retrieve the mean face of the class to which this VA value belongs. We use this mean face as the affect to be added on the 3D face reconstructed from the provided neutral image. Fig. \ref{framework} shows the proposed approach for this specific case. Fig. \ref{framework2} illustrates the procedure described in \ref{va_mean_face} given that the 550 VA classes are already created.

%+ add giati mporeis na kaneis synthesize temporal => logw vasis uparxei temporal consistency

\noindent \textbf{Temporal synthesis} If the user selects 'temporal synthesis', then, (s)he should provide a path  in the VA space (for instance by drawing) that the synthesized sequence should follow. Then, we retrieve the mean faces of the classes to which the VA values of the path belong. We use each of these mean faces as the affect to be added on the 3D faces reconstructed from the provided neutral image.
As a consequence, an expressive sequence is generated that shows the evolution of affect on the VA path specified by the user. 

%\textcolor{red}{** to keimenaki gia temporal: Here let us mention that the fact that the 4DFAB used in our approach is a temporal database, ensures that successive video frames' annotations are adjacent in the VA Space. Consequently, we generate temporal affect sequences from a given neutral face by utilizing 3D meshes that are adjacent in the VA Space, when computing the mean face of each class.}  

Here let us mention that the fact that the 4DFAB used in our approach is a temporal database, ensures that successive video frames' annotations are adjacent in the VA Space, since they generally show the same or slightly different states of affect. Thus, the 3D meshes of successive video frames will lie in the same and in adjacent classes in the 2-D VA space. Thus mean faces from adjacent classes can be used to show temporal evolution of affect as was above described.

%This provides our approach with the ability to select a sequence of mean faces from adjacent classes and synthesize temporal affect on the provided neutral face, following a path that the user provides (e.g., draws).

\subsubsection{Expression Blendshape Models}

Expression blendshape models provide an effective way to parameterize facial behaviors. The localized blendshape model~\cite{neumann2013sparse} has been used  to describe the selected VA samples. To build this model, we first bring all meshes into full correspondence following the dense registration approach described in Section~\ref{dense_corr}. As a result, we have a set of training meshes with the same number of vertices %(53215 vertices) 
and identical topology. Note that we have also selected one neutral mesh for each subject, which should have full correspondence with the rest data. Next, we subtract each 3D mesh from the respective neutral mesh, and create a set of $m$ difference vectors $\mathbf{d}_i \in \mathbb{R}^{3N}$. We then stack them into a matrix $\mathbf{D}=[\mathbf{d}_1, ..., \mathbf{d}_m] \in \mathbb{R}^{3N \times m}$, where $N$ is number of vertices in the mesh. Finally, a variant of sparse Principal Component Analysis (PCA) is applied to the data matrix $\mathbf{D}$, so as to identify sparse deformation components $\mathbf{C} \in \mathbb{R}^{h \times 1}$: 

\begin{equation}
    \arg \min \left \| \mathbf{D} - \mathbf{B} \mathbf{C} \right \|_F^2 + \Omega \left ( \mathbf{C} \right ) \; \; \; \textup{s.t.} \; \mathcal{V}\left (  \mathbf{B} \right ), 
\label{eq:blendshape_spca}
\end{equation}

\noindent where, the constraint $\mathcal{V}$ can be either $\max \left ( \left | \mathbf{B}_{k} \right | \right ) = 1, \; \forall k$ or $\max \left ( \mathbf{B}_{k} \right ) = 1, \; \mathbf{B} \geq 1, \; \forall k$, with $\mathbf{B}_k \in \mathbb{R}^{3N \times 1}$ denoting the $k^{th}$ components of sparse weight matrix $\mathbf{B} = [ \mathbf{B}_1, \cdots, \mathbf{B}_h]$. Selection of these two constraints depends on the actual usage; the major difference is that the latter one allows for negative weights and therefore enables deformation towards both directions, which is useful for describing shapes like muscle bulges. In this paper, we have selected the latter constraint, as we wish to enable bidirectional muscle movement and synthesise a rich variety of expressions. The regularization of sparse components $\mathbf{C}$ was performed with $\ell1 / \ell2$ norm~\cite{Wright2009,Bach2012}. To permit more local deformations, additional regularization parameters were added into $\Omega \left ( \mathbf{C} \right )$. To compute optimal $\mathbf{C}$ and $\mathbf{B}$, an iterative alternating optimization was employed (please refer to~\cite{neumann2013sparse} for more details).

\section{Databases}\label{databases}

To evaluate our facial affect synthesis method in different scenarios (e.g. controlled laboratory environment, uncontrolled in-the-wild setting), we utilized neutral facial images from as many as 13 databases (both small and large in terms of size). Table \ref{db_syn} briefly presents the Multi-PIE~\cite{gross2010multi}, Aff-Wild~\cite{kollias2019deep,zafeiriou2017aff}, AFEW 5.0~\cite{dhall2017individual}, AFEW-VA~\cite{kossaifi2017afew}, BU-3DFE~\cite{yin20063D}, RECOLA~\cite{ringeval2013introducing}, AffectNet~\cite{mollahosseini2017affectnet}, RAF-DB~\cite{li2017reliable}, KF-ITW~\cite{booth2017itw3dmm}, Face place, FEI~\cite{thomaz2010new}, 2D Face Sets and Bosphorus~\cite{savran2008bosphorus} databases that we used in our experimental study. Let us note that for AffectNet no test set is released and thus we use the released validation set to test on and randomly divide the training set into a training and a validation subset (with a 85/15 split).

Table \ref{db_syn} presents these databases by  showing: i) the model of affect they use, their condition, their type (static images or  audiovisual image sequences), the total number of frames and (male/female) subjects that they contain and the range of ages of the subjects, and ii) the total number of images that we synthesized using our approach (both in the valence-arousal and the six basic expressions cases).

\begin{table*}
\centering
\caption{Databases used in our approach, along with their properties and  the number of synthesized images in the valence-arousal case and the six basic expressions one;  'static' means images, 'A/V' means audiovisual sequences, i.e., videos }
\label{db_syn}
\scalebox{0.77}{
\begin{tabular}{ |c|c|c|c|c|c|c|c|c|c|  }
\hline 
Databases (DBs) & DB Type & Model of Affect & Condition & DB Size & \# of Subjects & Age Range & \multicolumn{2}{c|}{\begin{tabular}{@{}c@{}} Total \# of \\ Synthesized Images \end{tabular}} \\
\hline
& & & & & &  &  VA & Basic Expr \\
\hhline{|=|=|=|=|=|=|=|=|=|} 
MULTI-PIE~\cite{gross2010multi} & static & \begin{tabular}{@{}c@{}} Neutral, Surprise, Disgust, \\ Smile + Squint, Scream \end{tabular} & controlled &755,370 & \begin{tabular}{@{}c@{}} 337 \\ Male: 235 \\ Female: 102 \end{tabular} & - & 52,254 & 5,520 \\
\hline
Kinect Fusion ITW~\cite{booth2017itw3dmm} & static & Neutral, Happiness , Surprise & in-the-wild   & 3,264 & 17 & - & 116,235 & 12,236 \\
\hline
FEI~\cite{thomaz2010new} & static & Neutral, Smile & controlled  & 2,800 & \begin{tabular}{@{}c@{}} 200 \\ Male: 100 \\ Female: 100 \end{tabular} & 19-40  & 11,400 & 1,200 \\
\hline
Face place\footnote{Stimulus images courtesy of Michael J. Tarr, Center for the Neural Basis of Cognition and Department of Psychology, Carnegie Mellon University, http://www.tarrlab.org/} & static  & 6 Basic Expr, Neutral, Confusion & controlled & 6,574 & \begin{tabular}{@{}c@{}} 235 \\ Male: 143 \\ Female: 92 \end{tabular} & - &59,736 & 6,288 \\
\hline
AFEW 5.0~\cite{dhall2017individual} & A/V & 6 Basic Expr, Neutral & in-the-wild & 41,406 & $>$330 & 1-77 &  705,649 & 56,514  \\
\hline
RECOLA\cite{ringeval2013introducing} & A/V & VA & controlled  & 345,000  &  \begin{tabular}{@{}c@{}} 46 \\ Male: 19 \\ Female: 27 \end{tabular} & -  & 46,455  & 4,890 \\
\hline
BU-3DFE~\cite{yin20063D} & static  & 6 Basic Expr, Neutral  & controlled  & 2,500 & \begin{tabular}{@{}c@{}} 100 \\ Male: 56 \\ Female: 44 \end{tabular} & 18-70  & 5,700  & 600 \\
\hline
Bosphorus\cite{savran2008bosphorus} & static & 6 Basic Expr & controlled  &  4,666 & \begin{tabular}{@{}c@{}} 105 \\ Male: 60 \\ Female: 45 \end{tabular} & 25-35  & 17,018  & 1,792 \\
\hline
AffectNet\cite{mollahosseini2017affectnet} & static & \begin{tabular}{@{}c@{}} VA + 6 Basic Expr, \\  Neutral + Contempt \end{tabular} & in-the-wild  & \begin{tabular}{@{}c@{}}450,000 \\ manually \\ annotated \end{tabular} & - & 0 to $>$50  & 2,476,235  & 176,425 \\
\hline
Aff-Wild~\cite{kollias2019deep}~\cite{zafeiriou2017aff} & A/V & VA & in-the-wild  & 1,224,094 & \begin{tabular}{@{}c@{}} 200 \\ Male: 130 \\ Female: 70 \end{tabular} & -   & 60,135  & 6,330 \\
\hline
AFEW-VA~\cite{kossaifi2017afew} & A/V & VA & in-the-wild  & 30,050 & $<$600 & - & 108,864  & 11,460 \\
\hline
RAF-DB\cite{li2017reliable} & static & \begin{tabular}{@{}c@{}} 6 Basic,  Neutral \\ + 11 Compound Expr \end{tabular}  & in-the-wild  & \begin{tabular}{@{}c@{}} 15,339 \\ + 3,954 \end{tabular} & - & 0-70  & 121,866 & 12,828  \\
\hline
\begin{tabular}{@{}c@{}}  2D Face Sets\footnote{http://pics.stir.ac.uk}: \\ Pain \end{tabular} & static & \begin{tabular}{@{}c@{}} 6 Basic, Neutral \\ + 10 Pain Expr \end{tabular} & controlled   &  599 & \begin{tabular}{@{}c@{}} 23 \\ Male: 13 \\ Female: 10 \end{tabular} &- & 2,736  & 288  \\
\hline
\begin{tabular}{@{}c@{}}  2D Face Sets: \\ Iranian \end{tabular} & static & Neutral, Smile & controlled  & 369 & \begin{tabular}{@{}c@{}} 34 \\ Male: 0 \\ Female: 34 \end{tabular} &-   & 2,679  & 282  \\
\hline
\begin{tabular}{@{}c@{}}  2D Face Sets: \\ Nottingham Scans \end{tabular} & static & Neutral & controlled  & 100 & \begin{tabular}{@{}c@{}} 100 \\ Male: 50 \\ Female: 50 \end{tabular}  &-   & 5,700  & 600  \\
\hline
\end{tabular}
}
\end{table*}

\section{Experimental Study}

This section describes the experiments performed so as to evaluate the proposed approach. At first, we provide a qualitative evaluation of our approach by showing many synthesized images or image sequences from all thirteen databases described in the previous Section; as well as by comparing images generated by state-of-the-art GANs (StarGAN, Ganimation) and our approach. Next, a quantitative evaluation is performed by using the synthesized images as additional data to train Deep Neural Networks (DNNs); it is shown that the trained DNNs outperform current state-of-the-art networks and GAN-based methods on each database. Finally an ablation study is performed in which: i) the synthesized data are considered and used as a training (test) dataset, while the original data are respectively used as test (training) dataset, ii) the effect of the amount of synthesized data on network performance is studied, iii) an analysis is performed based on subjects' age.

\subsection{Qualitative evaluation of achieved facial affect synthesis}

We used all databases mentioned in Section \ref{databases} to supply the proposed approach with 'input' neutral faces. We then synthesized the emotional state corresponding to specific affects (both in VA case and in the six basic expressions one) for these images. At first we show many generated images (static synthesis) according to different VA values, then we illustrate examples of generated image sequences (temporal synthesis) and next we present some synthesized (static) images according to the six basic expressions. Finally, we visually compare images generated by our approach with synthesized images by StarGAN and Ganimation.

\subsubsection{Results on Static \& Temporal Affect Synthesis}

\begin{figure*}
\centering
\adjincludegraphics[width=1\linewidth]{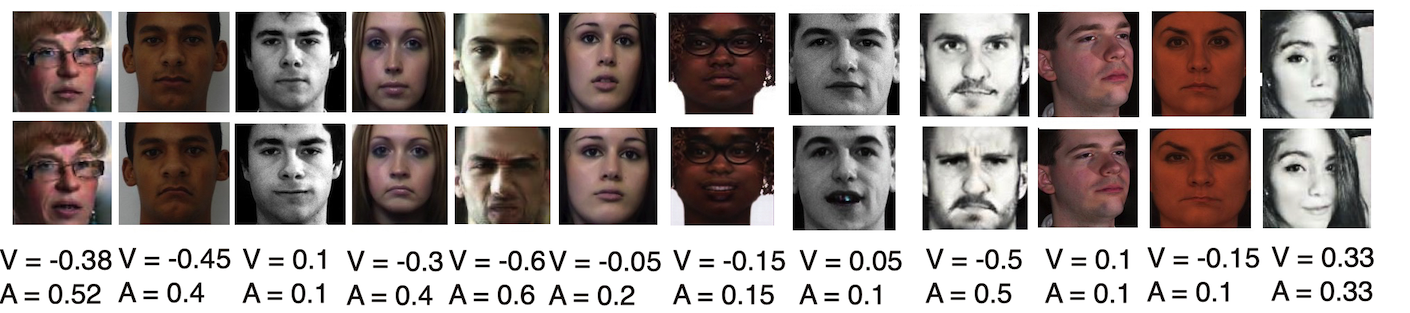}\\
(a) \\
%\vspace{5pt}
\adjincludegraphics[width=0.5\linewidth]{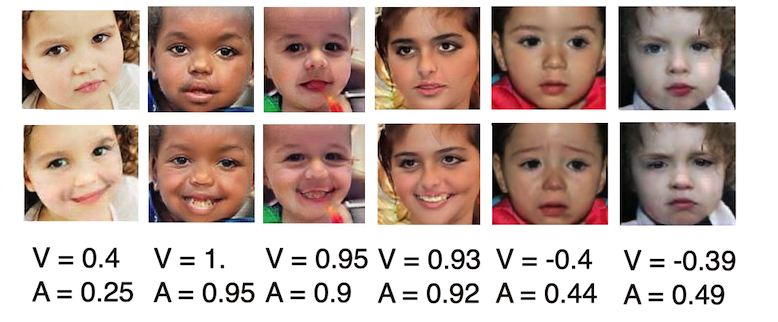}\adjincludegraphics[width=0.34\linewidth]{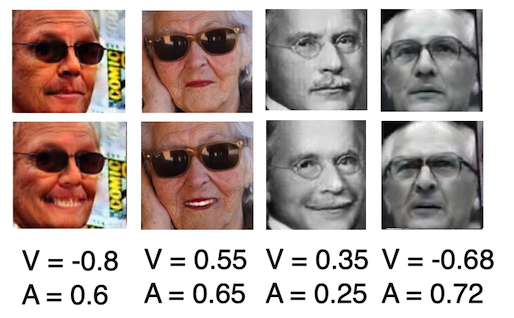}
\\
\HS \HS \HS \HS \HS \HS \HS \HS \HS \HS (b) \HS \HS \HS \HS \HS \HS \HS \HS \HS \HS \HS \HS \HS \HS \HS \HS \HS \HS \HS \HS \HS \HS \HS \HS \HS \HS \HS \HS \HS \HS \HS \HS \HS \HS \HS \HS \HS \HS \HS \HS  \HS \HS \HS \HS \HS \HS \HS \HS \HS \HS \HS \HS \HS \HS \HS \HS \HS \HS \HS \HS \HS \HS \HS \HS \HS \HS \HS \HS \HS (c) 
\\
\caption{\textbf{(a)-(c).} VA Case of static (facial) synthesis across all databases; first rows show the neutral, second ones show the corresponding synthesized images and third rows show the corresponding VA values. Images of: (b) kids, (c) elderly people and (a) in-between ages, are shown.}
\label{results_all_dbs}
\end{figure*}

\begin{figure}
\centering
\adjincludegraphics[height=14cm,width=1\linewidth]{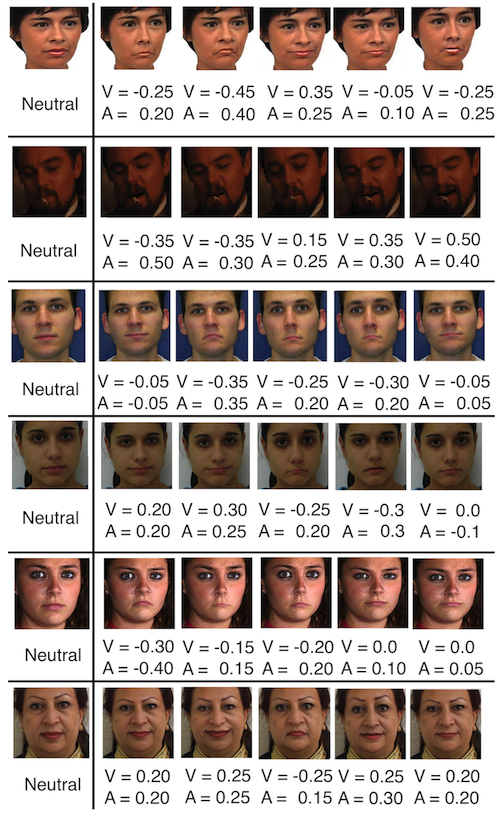} 
\caption{VA case of facial synthesis: on the left hand side are the neutral 2D images and on the right the synthesized images with different levels of affect}
\label{diff_expr}
\end{figure}

\begin{figure}[h]
\centering
\adjincludegraphics[height=3.2cm]{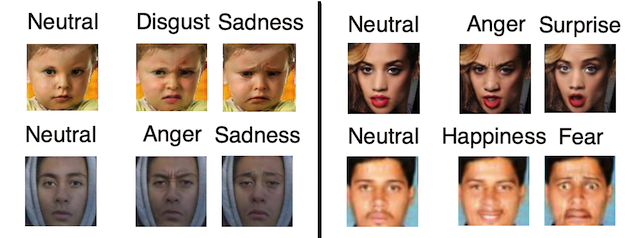} \\
\HS  \HS (a) \HS \HS \HS \HS \HS \HS  \HS \HS \HS \HS \HS  \HS \HS \HS \HS \HS \HS  \HS \HS \HS \HS \HS  \HS \HS \HS \HS \HS \HS  \HS \HS \HS \HS \HS \HS  \HS (b)
\caption{Basic Expression Case of facial synthesis: on the left hand side of (a) and (b) are the neutral 2D images and on the right the synthesized images with some basic expressions}
\label{basic_expr_exaple}
\end{figure}

\begin{figure*}
\centering
\adjincludegraphics[width=1\linewidth]{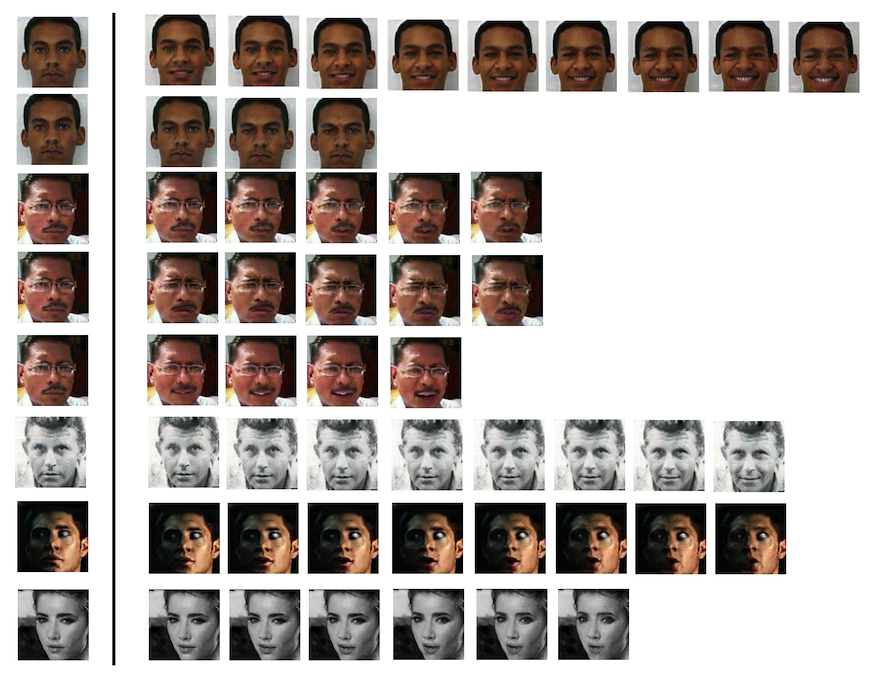} 
\caption{VA Case of temporal (facial) synthesis: on the left hand side are the neutral 2D images and on the right the synthesized image sequences}
\label{temporal}
\end{figure*}

Fig. \ref{results_all_dbs} shows representative results of facial affect synthesis, when user inputs a VA pair and selects to generate a static image. These results are organized in three age groups: Fig. \ref{results_all_dbs}(b) kids, Fig. \ref{results_all_dbs}(c) elderly people and Fig. \ref{results_all_dbs}(a) in-between ages. In each part, the first row illustrates neutral images sampled from each of the aforementioned databases, the second one shows the respective synthesized images and the third shows the respective VA values that were synthesized. Moreover, Fig. \ref{diff_expr} shows neutral images on the left hand side (first column) and  synthesized images, with various valence and arousal values, on the right hand side (following columns). It can be observed that the synthesized images are identity preserving, realistic and vivid. Fig. \ref{basic_expr_exaple} refers to the basic expression case; it shows neutral images on the left hand side of (a) and (b) and  synthesized images with basic expressions on the right hand side. Fig. \ref{temporal} illustrates the VA case for temporal synthesis, as was described in Section 4.5.2. Neutral images are shown on the left hand side, while synthesized face sequences with time-varying levels of affect are shown on the right hand side.

All these Figs. show that the proposed framework works well, when using images from either in-the-wild, or controlled databases. This indicates that we can effectively synthesize facial affect irregardless of image conditions (e.g., occlusions, illumination and head poses). 

\subsubsection{Comparison with GANs}\label{comp-qual}

In order to characterize the value that the proposed approach imparts, we provide qualitative comparisons with two state-of-the-art GANs, namely StarGAN \cite{choi2018stargan} and Ganimation. Like CycleGAN (referenced in Section \ref{related_work}), Star-GAN performs image-to-image translation, but adopts a unified approach such that a single generator is trained to map an input image to one of multiple target domains, selected by the user. By sharing the generator weights among different domains, a dramatic reduction of the number of parameters is achieved. Ganimation was described in Section \ref{related_work}.

We used these networks to fit the VA case, namely to generate expressions according to VA values.  We trained them with the same 600,000 frames of 4DFAB that we used in our approach. Preprocessing also included face detection and alignment. For a fair comparison, in all presented comparisons (both qualitative and quantitative), the GANs generated samples, while provided with the same neutral images and the same VA values.

\begin{figure*}
\centering
\adjincludegraphics[width=1\linewidth]{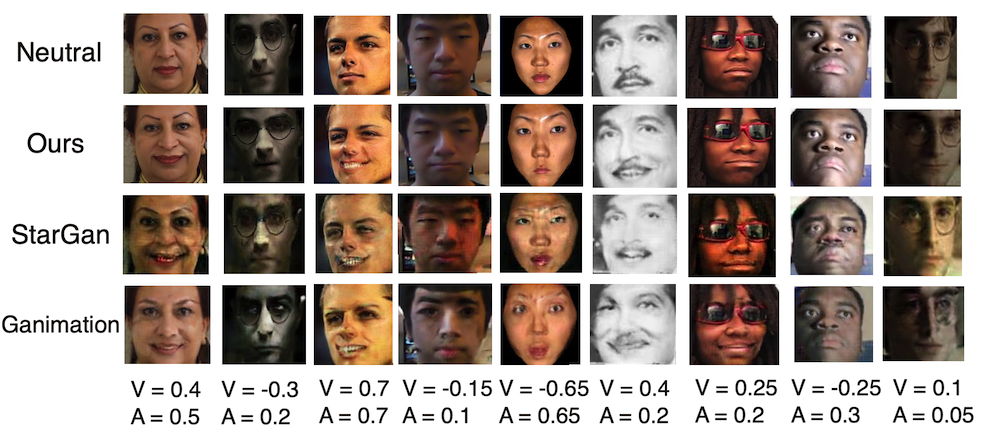} \\
\adjincludegraphics[width=1\linewidth]{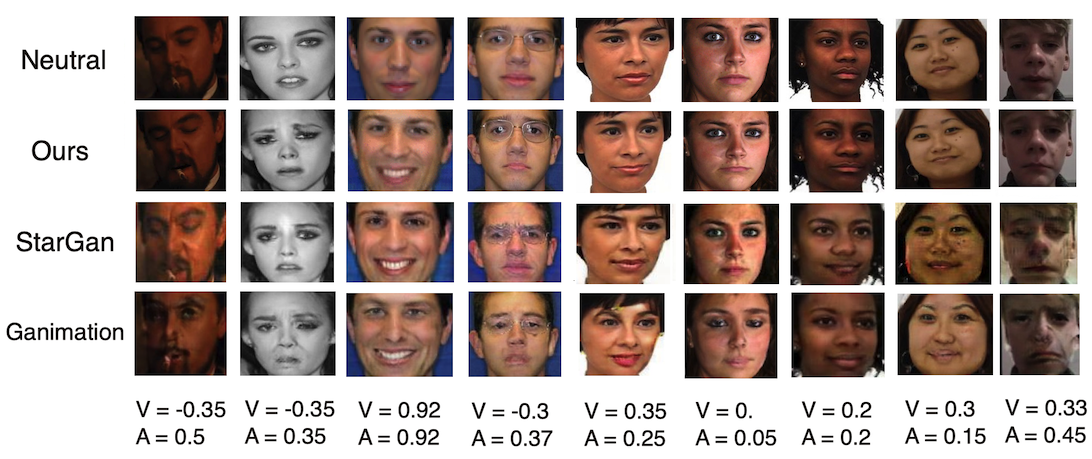} 
\caption{Generated results by our approach, StarGAN and Ganimation}
\label{compare}
\end{figure*}

Fig. \ref{compare} presents a visual comparison between images generated by our approach, StarGAN and Ganimation. It shows the neutral images, the synthesized VA values and the resulting images. It is evident that our approach synthesizes samples that: i) look much more natural and realistic, ii) maintain the degree of sharpness of the original neutral image, and iii) combine visual accuracy with spatial resolution. 

Some further deductions can be made from Fig. \ref{compare}. 
StarGAN does not perform well when tested on different in-the-wild and controlled databases that include variations in illumination conditions and head poses. StarGAN is shown to not reflect detailed illumination; unnatural lighting changes were observed on the results. These can be explained because in the original StartGAN paper \cite{choi2018stargan}, its capability to generate affect has not been tested on in-the-wild facial analysis (we refer only to the case of emotion recognition). In general, StarGAN yields more realistic results when it is trained simultaneously with multiple datasets annotated for different tasks. 

Additionally, in \cite{choi2018stargan}, when referring to emotion recognition, StarGAN was trained and evaluated on Radboud Faces Database (RaFD) \cite{langner2010presentation} which: i) is very small in terms of size (around 4,800 images) and ii) is a lab-controlled and posed expression database. Last but not least, StarGAN has been tested to change only a particular aspect of a face among a discrete number of attributes/emotions defined by the annotation granularity of the dataset. As can be seen in Fig. \ref{compare}, StarGAN cannot accurately provide realistic results when tested in the much broader and more difficult task of valence and arousal generation (and estimation).

As far as Ganimation is concerned, its results are also worse than the results of our approach. In most cases, it shows artifacts and in some cases certain levels of blurriness. When compared to StarGAN, Ganimation seems more robust to changing backgrounds and lighting conditions; this is due to the attention and color masks that it contains. Nevertheless, in general, errors in the attention mechanism occur when the input contains extreme expressions. The attention mechanism does not seem to  sufficiently weight the color transformation, causing transparencies. It is interesting to note that on the Leonardo DiCaprio image, the synthesized image by Ganimation shows open eyes, whereas on the neutral image (and the one synthesized by our approach) eyes are closed; this illustrates errors of the mask. %\textcolor{red}{Presumably the attention mask could be the reason why the produced images are not realistic enough. The attention mask may be good when dealing with actions of some specific facial muscles (14 AUs) but it shows to not be as good to cope well with the valence and arousal case. The 2D VA Space contains the whole spectrum of emotions and is not shown to be directly related with some specific muscles (whereas action units and basic expressions are directly related; see for example \cite{du2014compound}). All facial muscles are related to the VA problem.} 
For example, in Fig. \ref{compare}, images produced by Ganimation in columns 1, 3, 4, 5, 6, 9 show the discussed problems.

\subsection{Quantitative evaluation of the facial affect synthesis through data augmentation}

It is generally accepted that using more training data - of good quality - leads
to better results in supervised training. Data augmentation increases the effective size of the training dataset.
In this Section we present a data augmentation strategy which uses the synthesized data produced by our approach, as additional data to train DNNs, for both valence-arousal prediction, as well as classification into the basic expression categories. In particular, we describe experiments performed on eight databases, presenting the adopted evaluation criteria, the networks we used and the obtained results. We also report the performances of the networks trained -in a data augmentation manner- with synthesized images from StarGAN and Ganimation. It is shown that the DNNs trained with the proposed data augmentation methodology outperform both the state-of-the-art techniques and the DNNs trained with StarGAN and Ganimation, in all experiments, validating the effectiveness of the proposed facial synthesis approach. Let us first explain some notations. In the followings, by reporting 'network\_name trained using StarGAN', 'network\_name trained using Ganimation' and 'network\_name trained using the proposed approach', we refer to networks trai-ned with the specific database's training set augmented with data synthesized by StarGAN, Ganimation and the proposed approach, respectively.

\subsubsection{Leveraging synthesized data for training Deep Neural Networks: Valence-Arousal case}\label{va_augm}

In this set of experiments we consider four facial affect databases annotated in terms of valence and arousal, the Aff-Wild, RECOLA, AffectNet and AFEW-VA data-bases. 
At first, we selected neutral frames from these databases, i.e., frames with zero valence and arousal values (human inspection was also conducted to make sure that they represented neutral faces). For every frame, we synthesized facial affect according to the methodology described in Section \ref{methodology}. We start by first describing the evaluation criteria used in our experiments.

\paragraph{The adopted evaluation criteria}

The main evaluation criterion that we use is the Concordance Correlation Coefficient (CCC) \cite{lawrence1989concordance}, which has been widely used in related Challenges (e.g., \cite{valstar2016avec}); we also report the Mean Squared Error (MSE), since this has been also frequently used in related research.

CCC evaluates the agreement between two time series by scaling their correlation coefficient with their mean square difference. CCC takes values in the range $[-1,1]$, where $+1$ indicates perfect concordance and $-1$ denotes perfect discordance. Therefore high values are desired.
CCC is defined as follows:
\begin{equation} \label{eq:6}
\rho_c = \frac{2 s_{xy}}{s_x^2 + s_y^2 + (\bar{x} - \bar{y})^2},
\end{equation}
where $s_x$ and $s_y$ are the variances of the ground truth and predicted values respectively, $\bar{x}$ and $\bar{y}$ are the corresponding mean values and $s_{xy}$ is the respective covariance value.

The Mean Squared Error (MSE) provides a simple comparative metric, with a small value being desirable. MSE is defined as follows:
\begin{equation} \label{mse}
MSE = \frac{1}{N} \sum_{i=1}^{N} (x_i-y_i)^2,
\end{equation}
where $x$ and $y$ are the ground truth and predicted values respectively and $N$ is the total number of samples.

In some cases we also report the Pearson-CC (P-CC) and the Sign Agreement Metric (SAGR), since they have been reported by respective state-of-the-art methods.

The P-CC takes values in the range [-1,1] and high values are desired. It is defined as follows:
\begin{equation} \label{eq:1}
\rho_{xy} = \frac{s_{xy}}{s_x  s_y},
\end{equation}
where $s_x$ and $s_y$ are the variances of the ground truth and predicted values respectively and $s_{xy}$ is the respective covariance value.

The SAGR takes values in the range [0,1], with high values being desirable. It is defined as follows:
\begin{equation} \label{eq:10}
SAGR = \frac{1}{N} \sum_{n=1}^{N} \delta(sign(x_i),sign(y_i)),
\end{equation}
where $N$ is the total number of samples, $x$ and $y$ are the ground truth and predicted values respectively, $\delta$ is the Kronecker delta function and $\delta(sign(x),sign(y))$ is defined as:

\begin{equation} \label{eq:11}
\delta(sign(x),sign(y)) = 
     \begin{cases}
       \text{1,} &\quad\text{$x \geqslant 0$ and $y \geqslant 0$}\\
       \text{1,} &\quad\text{$x \leqslant 0$ and $y \leqslant 0$}\\
       \text{0,} &\quad\text{otherwise} \\
     \end{cases}
\end{equation}

\paragraph{Experiments on Dimensional Affect} \label{dimensional}

\textbf{Aff-Wild} We synthesized 60,135 images from the Aff-Wild database and added those images to the training set of the first Affect-in-the-wild Challenge.
The employed network architecture was the AffWildNet (VGG-FACE-GRU) described in \cite{kollias2017recognition,kollias2019deep}.

Table \ref{affwild_db} shows a comparison of the performance of: the VGG-FACE-GRU  trained using: i) our approach, ii) StarGAN, and iii) Ganimation; the best performing network, AffWildNet, reported in \cite{kollias2017recognition,kollias2019deep}; the winner of the Aff-Wild Challenge \cite{weichi} (FATAUVA-Net).

\begin{table}[h]
\caption{Aff-Wild: CCC and MSE evaluation of valence \& arousal predictions provided by the VGG-FACE-GRU trained using our approach vs state-of-the-art networks and methods. Valence and arousal values are in $[-1,1$]. }
\label{affwild_db}
\centering
\scalebox{0.83}{
\begin{tabular}{ |c||c|c|c|c| }
 \hline
 \multicolumn{1}{|c||}{Networks} & \multicolumn{2}{c|}{CCC} & \multicolumn{2}{c|}{MSE}   \\
 \hline
     & Valence & Arousal & Valence & Arousal  \\
 \hline
FATAUVA-Net \cite{weichi}  & 0.396 & 0.282 & 0.123 & 0.095 		   \\
 \hline
\begin{tabular}{@{}c@{}}VGG-FACE-GRU \\ trained using StarGAN \end{tabular} & 0.556 & 0.424  & 0.085  & 0.060  \\
 \hline  
\begin{tabular}{@{}c@{}}VGG-FACE-GRU \\ trained using Ganimation \end{tabular}& 0.576  & 0.433  & 0.077  &  0.057	\\
\hline
AffWildNet \cite{kollias2017recognition,kollias2019deep} &0.570	&0.430 & 0.080 & 0.060	\\
 \hline
\begin{tabular}{@{}c@{}} VGG-FACE-GRU \\ trained using the \\ proposed approach \end{tabular}  & \textbf{0.595}&	\textbf{0.445} & \textbf{0.074} & \textbf{0.051}	\\
 \hline 
\end{tabular}
}
\end{table}

\begin{figure}[h!]
\centering
\adjincludegraphics[height=5cm]{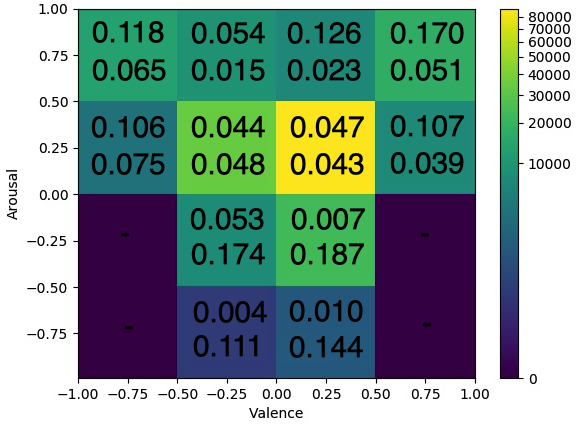}\\
(a)\\
\adjincludegraphics[height=5cm]{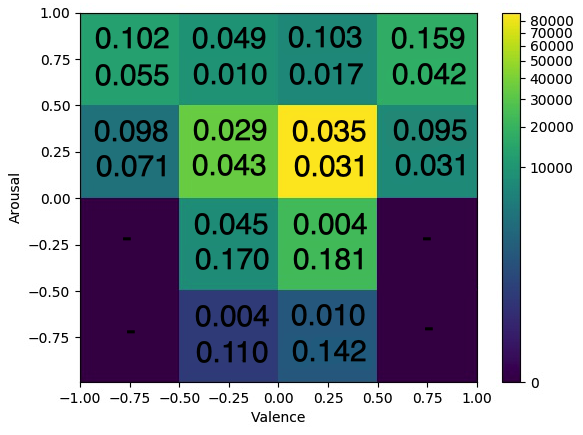}\\
(b)\\
\caption{The 2D histogram of valence and arousal Aff-Wild's test set annotations, along with the MSE per grid area, in the case of (a) AffWildNet and (b) VGG-FACE-GRU trained using the proposed approach
}
\label{mse_va_affwild}
\end{figure}

From Table \ref{affwild_db}, it can be verified that the network trained on the augmented dataset, with synthesized by our approach images, outperformed all other networks. It should be noted that the number of synthesized images (around $60K$) was small compared to the size of Aff-Wild's training set (around $1M$), the latter being already sufficient for training the best performing DNN; consequently, the improvement was not large, about 2\%. An interesting observation is that the network trained using StarGAN displayed worse performance than AffWildNet. This means that the 68 landmark points that were passed as additional input to the AffWildNet helped the network in reaching a better performance than just adding a small amount (compared to the training set size) of auxiliary synthesized data.
The MSE error improvement on Valence and Arousal estimation provided by the augmented training vs the AffWildNet one, over the different areas of the VA space, is shown through the 2D histograms presented in Fig. \ref{mse_va_affwild}. It can be seen that the improvement on MSE was better in areas in which a larger number of new samples was generated, i.e., in the positive valence regions.

\noindent \textbf{RECOLA}  We generated 46,455 images from RECOLA; this number corresponds to around 40\% of its training data set size. The employed network architecture was the ResNet-GRU described in \cite{kollias2019deep}.

Table \ref{recola_db} shows a comparison of the performance of: the ResNet-GRU network trained using: i) our approach, ii) StarGAN, and iii) Ganimation; the AffWildNet fine-tuned on the RECOLA, as reported in \cite{kollias2019deep}; a ResNet-GRU directly trained on RECOLA, as reported in \cite{kollias2019deep}.

\begin{table}[h]
\caption{RECOLA: CCC evaluation of valence \& arousal predictions provided by the ResNet-GRU trained using the proposed approach vs other state-of-the-art networks and methods.}
\label{recola_db}
\centering
\begin{tabular}{ |c||c|c| }
 \hline
 \multicolumn{1}{|c||}{Networks} & \multicolumn{2}{c|}{CCC}   \\
 \hline
     & Valence & Arousal   \\
 \hline
ResNet-GRU \cite{kollias2019deep}  & 0.462 & 0.209 		   \\
 \hline
\begin{tabular}{@{}c@{}}ResNet-GRU \\ trained using StarGAN \end{tabular}&  0.503 & 0.245 	\\
\hline
\begin{tabular}{@{}c@{}}ResNet-GRU \\  trained using Ganimation \end{tabular} & 0.486 & 0.222 \\
 \hline  
fine-tuned AffWildNet\cite{kollias2019deep} &0.526	&0.273 	\\
 \hline
\begin{tabular}{@{}c@{}}ResNet-GRU trained \\ using the proposed approach \end{tabular} & \textbf{0.554}&	\textbf{0.312} 	\\
 \hline 
\end{tabular}
\end{table}

\begin{figure}[h!]
\centering
\adjincludegraphics[height=5.5cm]{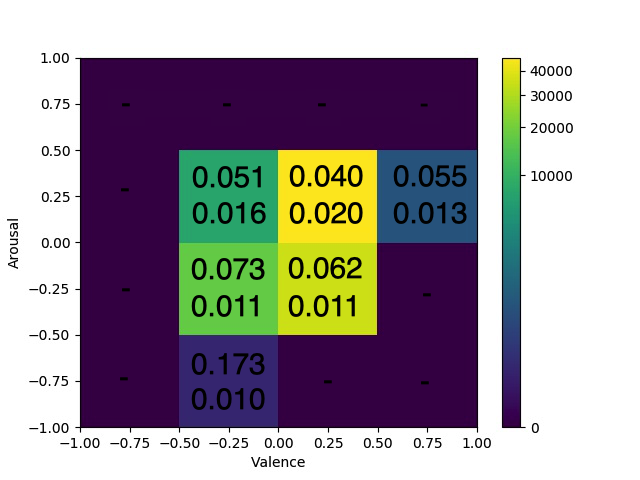}\\
(a)\\
\adjincludegraphics[height=5cm,trim={0.1\totalheight} {0\totalheight} {0\totalheight} {0\totalheight}]{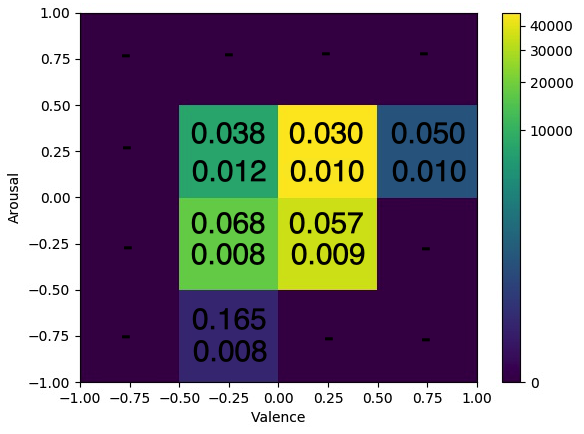}\\
(b)\\
\caption{The 2D histogram of valence and arousal RECOLA's test set annotations, along with the MSE per grid area, in the case of (a) ResNet-GRU and (b) ResNet-GRU trained using the proposed approach}
\label{mse_va_recola}
\end{figure}

\begin{table*}[h]
\caption{AffectNet: CCC, P-CC, SAGR and MSE  evaluation of valence \& arousal predictions provided by the VGG-FACE trained using the proposed approach vs  state-of-the-art networks and methods. Valence and arousal values are in $[-1,1$]. }
\label{affectnet_db}
\centering
\begin{tabular}{ |c||c|c|c|c|c|c|c|c| }
 \hline
 \multicolumn{1}{|c||}{Networks} & \multicolumn{2}{c|}{CCC} & \multicolumn{2}{c|}{P-CC} & \multicolumn{2}{c|}{SAGR}  & \multicolumn{2}{c|}{MSE}\\
 \hline
     & Valence & Arousal  & Valence & Arousal  & Valence & Arousal & Valence & Arousal  \\
 \hline
AlexNet \cite{mollahosseini2017affectnet}  & 0.60 & 0.34 & \textbf{0.66} & 0.54 & 0.74 & 0.65 & \textbf{0.14} & 0.17		   \\
 \hline
 \begin{tabular}{@{}c@{}}the VGG-FACE baseline \end{tabular} & 0.50 & 0.37 & 0.54 & 0.48  & 0.65 & 0.60  & 0.19  &	0.18 	\\
 \hline 
\begin{tabular}{@{}c@{}}VGG-FACE \\ trained using StarGAN  \end{tabular} & 0.55 & 0.42 & 0.58 & 0.49 & 0.74 & 0.73 & 0.17 & 0.16	 	\\
 \hline  
\begin{tabular}{@{}c@{}}VGG-FACE trained \\ using Ganimation \end{tabular} & 0.56 & 0.45 & 0.59 & 0.51 & 0.74 & 0.74 & 0.15 & 0.16	 	\\
 \hline   
\begin{tabular}{@{}c@{}}VGG-FACE trained \\ using the proposed approach \end{tabular} & \textbf{0.62} & \textbf{0.54} & \textbf{0.66} & \textbf{0.55} & \textbf{0.78} & \textbf{0.75} & \textbf{0.14} & \textbf{0.15}	 	\\
 \hline 
\end{tabular}
\end{table*}

From Table \ref{recola_db}, it can be verified that the network trained using the proposed approach outperformed all other networks. The above gains in performance can be justified by the fact that the number of synthesized images (around 46,500) was significant compared to the size of RECOLA's training set (around 120,000) and that the original training set size was not very sufficient to train the DNNs. It is worth mentioning that the GAN based methods have not managed to provide a sufficiently enriched dataset so that a similar boost in the achieved performances could be obtained. The MSE error improvement on Valence and Arousal estimation provided by the augmented training vs the original one (which was  0.045-0.100 vs 0.055-0.160), over the different areas of the VA space, is shown through the 2D histograms presented in Fig. \ref{mse_va_recola}. Big reduction of MSE value was achieved in all covered VA areas.

\noindent \textbf{AffectNet} The AffectNet database contains around 450,000 manually annotated images and around 550,000 automatically annotated images for valence-arousal. We only used the manually annotated images so as to be consistent with the state-of-the-art networks that were also trained using this set. Additionally, the manually annotated set ensures that the images used by our approach to synthesize new, are indeed neutral.  We created 2,476,235 synthesized images from the AffectNet database, a number that is more than 5 times bigger than the training data size. The employed network architecture was VGG-FACE. For comparison purposes, we trained the network using the original training data set (let us call this network 'the VGG-FACE baseline').

Table \ref{affectnet_db} shows a comparison of the performance of: the VGG-FACE baseline; the VGG-FACE  trained using: i) our approach, ii) StarGAN, and iii) Ganimation; AlexNet, which is the baseline network of the AffectNet database \cite{mollahosseini2017affectnet}.

\begin{figure}[h!]
\centering
\adjincludegraphics[height=5cm]{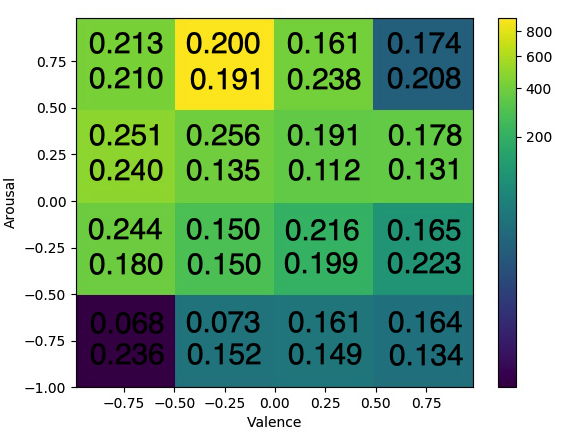}\\
(a)\\
\adjincludegraphics[height=5cm]{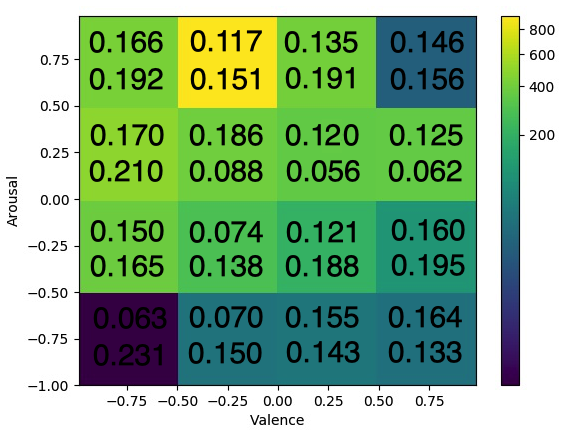}\\
(b)\\
\caption{The 2D histogram of valence and arousal AffectNet's test set annotations, along with the MSE per grid area, in the case of (a) the VGG-FACE baseline, (b) the VGG-FACE trained using the proposed approach}
\label{mse_va_affectnet}
\end{figure}

\begin{figure}[h!]
\centering
\adjincludegraphics[height=6cm]{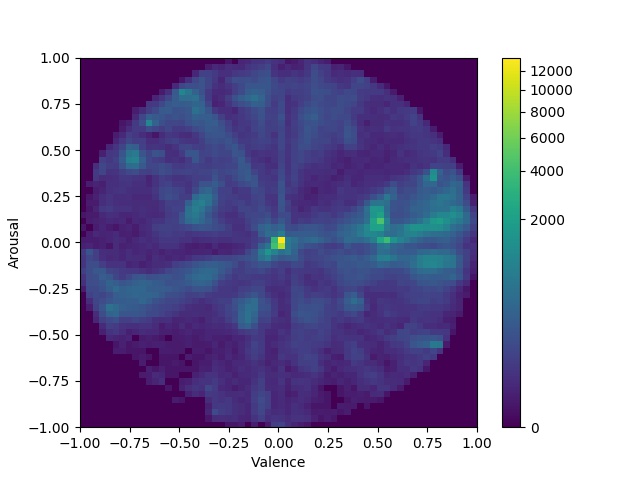}\\
\caption{The 2D histogram of valence and arousal AffectNet's annotations for the manually annotated training set}
\label{hist_va_affectnet}
\end{figure}

From Table \ref{affectnet_db}, it can be verified that the network trained by the proposed methodology outperformed all other networks. This boost in performance has been large, in all evaluation criteria, compared to the VGG-FACE baseline network, with spread of this improvement over the VA space shown in Fig. \ref{mse_va_affectnet}. The explanation arises from the large number of synthesized images that helped the network train and generalize better, since in the training set there existed a lot of ranges that were poorly represented. This is shown in the histogram of the -manually annotated- training set, for valence and arousal, in Fig. \ref{hist_va_affectnet}. Our network also outperformed the AffectNet's database baseline. For the arousal estimation, the performance gain was remarkable, mainly in CCC and SAGR evaluation criteria, whereas for the valence estimation the performance gain was also significant.

\noindent \textbf{AFEW-VA}. We synthesized 108,864  images from the AFEW-VA database, a number that is more than 3.5 times bigger than its original size. For training, we used the VGG-FACE-GRU architecture described in \cite{kollias2019deep}. Similarly to \cite{kossaifi2017afew}, we used a 5-fold person-independent cross-validation strategy and at each fold we augmented the training set with the synthesized images of people appearing only in that set (preserving the person independence).

Table \ref{afew_va_db} shows a comparison of the performance of: the VGG-FACE-GRU network trained  using: i) our approach, ii) StarGAN, and iii) Ganimation; the best performing  network as reported in \cite{kossaifi2017afew}.

\begin{table}[h]
\caption{AFEW-VA: P-CC and MSE  evaluation of valence \& arousal predictions provided by the VGG-FACE trained using the proposed approach vs state-of-the-art network and methods. Valence and arousal values are in $[-1,1$].}
\label{afew_va_db}
\centering
\scalebox{0.82}{
\begin{tabular}{ |c||c|c|c|c|  }
 \hline
 \multicolumn{1}{|c||}{Networks} & \multicolumn{2}{c|}{Pearson CC}  & \multicolumn{2}{c|}{MSE} \\
 \hline
     & Valence & Arousal & Valence & Arousal  \\
\hline
best of \cite{kossaifi2017afew} & 0.407 & 0.450 & 0.484 & 0.247 \\
\hline
\begin{tabular}{@{}c@{}}VGG-FACE \\ trained using StarGAN  \end{tabular}&  0.512 & 0.489 & 0.262 & 0.097	 	\\
 \hline
 \begin{tabular}{@{}c@{}}VGG-FACE \\ trained using Ganimation  \end{tabular} & 0.491 & 0.453 & 0.308 & 0.151	 	\\
 \hline  
\begin{tabular}{@{}c@{}} VGG-FACE-GRU \\ trained using \\ the proposed approach \end{tabular}  & \textbf{0.562} & \textbf{0.614}	& \textbf{0.226} &  \textbf{0.075} \\
\hline
\end{tabular}
}
\end{table}

\begin{figure}[h!]
\centering
\adjincludegraphics[height=5cm]{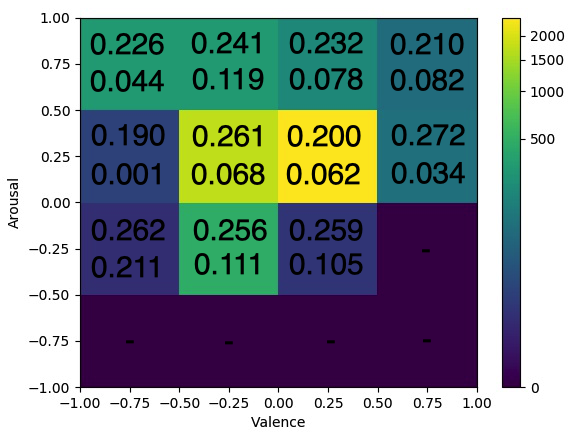}\\
\caption{The 2D histogram of valence and arousal AFEW-VA's test set annotations, along with the MSE per grid area, in the case of the  VGG-FACE trained using the proposed approach}  %; the results are the average of 5 fold cross validation}
\label{mse_va_afewva}
\end{figure}

From Table \ref{afew_va_db}, it can be verified that the network trained using the proposed approach outperformed all other networks. Great boost in performance was achieved. The general gain in performance can be justified by the fact that the number of synthesized images (around 109,000) is  much greater  than the number of images in the dataset (around 30,000), with the latter being rather small for effectively training the DNNs. 
The 2D histogram in Fig. \ref{mse_va_afewva} shows the achieved MSE when using the proposed approach over the different areas of the VA space. 

\subsubsection{Leveraging synthesized data for training Deep Neural Networks: Basic Expressions case}

In the following experiments we used the synthesized faces to train DNNs, for classification into the six basic expressions, over four facial affect databases, RAF-DB, AffectNet, AFEW and BU-3DFE. 
Our first step has been to select neutral frames from these four databases. Then, for each frame, we synthesized facial affect according to the methodology described in Section \ref{methodology}. We start by first describing the evaluation criteria used in our experiments.

\paragraph{The adopted evaluation criteria}

One evaluation criterion used in the experiments is
total accuracy, defined as the total number of correct predictions divided by the total number of samples. Another criterion is the $F_1$ score, which is a weighted average of the recall (= the ability of the classifier to find all the positive samples) and precision (= the ability of the classifier not to label as positive a sample that is negative). The $F_1$ score reaches its best value at 1 and its worst score at 0. In our multi-class problem, $F_1$ score is the unweighted mean of the $F_1$ scores of the expression classes. $F_1$ score of each class is defined as:

\begin{equation} \label{f1}
F_1 = \frac{2 \times precision \times recall}{precision + recall}
\end{equation}

Another criterion that is used is the average of the diagonal values of the confusion matrix for the seven basic expressions. 

One, or more of the above criteria are used in our experiments, so as to illustrate the comparison with other state-of-the-art methods.

\paragraph{Experiments on  Categorical Affect}

\begin{table*}[h]
\caption{RAF-DB: The diagonal values of the confusion matrix for the seven basic expressions and their average, using the VGG-FACE trained using the proposed approach, as well as using other state-of-the-art networks.}
\label{RAF-DB_db}
\centering
\begin{tabular}{ |c||c|c|c|c|c|c|c|c| }
 %\multicolumn{1}{|c||}{Networks} & \multicolumn{1}{|c}{Basic} \\
\hline
Networks   & Anger & Disgust  & Fear & Happy  & Sad & Surprise & Neutral & Average  \\
\hhline{|=|=|=|=|=|=|=|=|=|} 
 LDA-VGG-FACE \cite{li2017reliable} & 0.661 & 0.250 & 0.378 & 0.731 & 0.515 & 0.535 & 0.472 & 0.506		   \\
 \hline
mSVM-VGG-FACE \cite{li2017reliable}  & 0.685 & 0.275 & 0.351 & 0.853 & 0.649 & 0.663 & 0.599 & 0.582		   \\
 \hline
the VGG-FACE baseline  & 0.691 & 0.287 & 0.363 & 0.853 & 0.661 & 0.666 & 0.600 & 0.589		   \\
 \hline
 mSVM-DLP-CNN \cite{li2017reliable} & 0.716 & 0.522 & \textbf{0.622} & \textbf{0.928} & 0.801 & 0.812 & 0.803 & 0.742		   \\
 \hline
\begin{tabular}{@{}c@{}}VGG-FACE trained \\ using the proposed approach \end{tabular} & \textbf{0.784} & \textbf{0.644} & \textbf{0.622} & 0.911  & \textbf{0.812} & \textbf{0.845}  & \textbf{0.806}  & \textbf{0.775} 	\\
 \hline 
\end{tabular}
\end{table*}

\begin{figure*}[h!]
\centering
\adjincludegraphics[height=6cm]{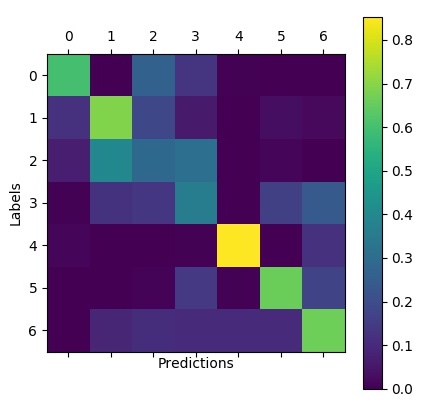}
 \HS \HS \HS \HS \HS 
 \adjincludegraphics[height=6cm]{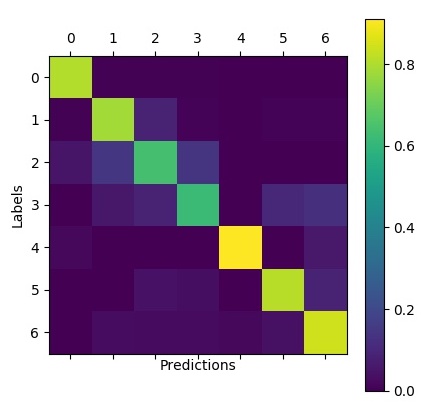}
 \\
  (a)\HS \HS \HS \HS \HS \HS \HS \HS \HS \HS \HS \HS \HS \HS \HS \HS \HS \HS \HS \HS \HS \HS \HS \HS \HS \HS \HS \HS \HS \HS \HS \HS \HS \HS \HS \HS \HS \HS \HS \HS \HS \HS \HS \HS \HS \HS \HS \HS \HS \HS \HS \HS \HS \HS \HS \HS \HS \HS \HS \HS \HS \HS \HS \HS \HS \HS \HS \HS (b)\\
\caption{The confusion matrix of (a) the VGG-FACE baseline and (b) the VGG-FACE trained using the proposed approach for the RAF-DB database; 0: Neutral, 1: Anger, 2: Disgust, 3: Fear, 4: Joy, 5: Sadness, 6: Surprise}
\label{conf_matrix_rafdb}
\end{figure*}

\textbf{RAF-DB}. In this database we only considered the six basic expression categories, since our approach synthesizes images based on these categories; we ignored compound expressions that were included in the original dataset. We created 12,828 synthesized images, which are slightly more than the training images (12,271). We employed the VGG-FACE network. For comparison purposes, we trained the network using the original training dataset (let us call this network ’the VGG-FACE baseline’).

For further comparison purposes, we used the networks defined in \cite{li2017reliable}: i) mSVM-VGG-FACE: first the VGG-FACE was trained on the RAF-DB database and then features from the penultimate fully connected layer were extracted and fed into a Support Vector Machine (SVM) that performed the classification, ii) LDA-VGG-FACE: same as before: LDA was applied on the features which were extracted from the penultimate fully connected layer and performed the final classification and iii) mSVM-DLP-CNN: the designed Deep Locality Preserving CNN network (we refer the interested reader for more details to \cite{li2017reliable}) was first trained on the RAF-DB database and then a SVM performed the classification using the features extracted from the penultimate fully connected layer of this architecture.

Table \ref{RAF-DB_db} shows a comparison of the performance of the above described networks. From Table \ref{RAF-DB_db}, it can be verified that the network trained using the proposed approach outperformed all state-of-the-art nets. When compared to the mSVM-VGG-FACE and LDA-VGG-FACE networks, the boost in performance has been significant. This can be explained by the fact that the disgust and fear classes, originally, did not contain a lot of training images, but after adding the synthesized data, they did. This resulted in obtaining a better performance in the other classes, as well. Interestingly, there was also a considerable performance gain in the neutral class, that did not contain any synthesized images. This can be explained by considering the fact that the network trained with the augmented data could distinguish better the classes, since it had more samples in the two above described categories. Fig. \ref{conf_matrix_rafdb} illustrates the whole confusion matrix of the VGG-FACE baseline and the VGG-FACE trained using the proposed approach, giving a better insight on the improved performance and verifying the above explanations.

\begin{figure*}[h!]
\centering
\adjincludegraphics[height=6cm]{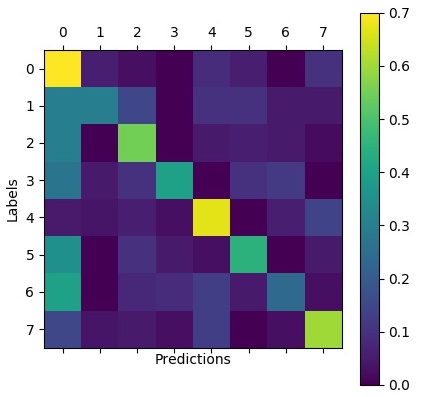}
 \HS \HS \HS \HS \HS 
 \adjincludegraphics[height=6cm]{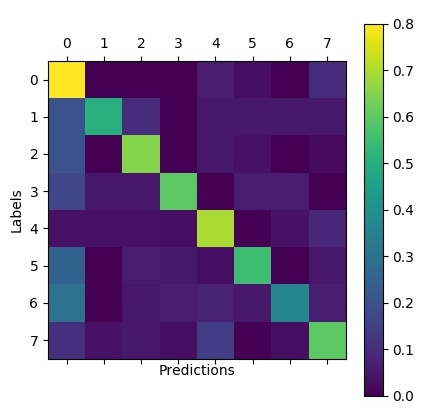}
 \\
  (a)\HS \HS \HS \HS \HS \HS \HS \HS \HS \HS \HS \HS \HS \HS \HS \HS \HS \HS \HS \HS \HS \HS \HS \HS \HS \HS \HS \HS \HS \HS \HS \HS \HS \HS \HS \HS \HS \HS \HS \HS \HS \HS \HS \HS \HS \HS \HS \HS \HS \HS \HS \HS \HS \HS \HS \HS \HS \HS \HS \HS \HS \HS \HS \HS \HS \HS \HS \HS (b)\\
\caption{The confusion matrix of (a) the VGG-FACE baseline and (b) the VGG-FACE trained using the proposed approach for the AffectNet database; 0: Neutral, 1: Anger, 2: Disgust, 3: Fear, 4: Joy, 5: Sadness, 6: Surprise, 7: Contempt}
\label{conf_matrix_affectnet}
\end{figure*}

\noindent \textbf{AffectNet}. We synthesized 176,425 images from the AffectNet database, a number that is almost 40\% of its size. It should be mentioned that the AffectNet database contained the six basic expressions and another one, contempt. Our approach synthesized images only for the  basic expressions, so for the contempt class we only kept the original training data. The network architecture that we employed here was VGG-FACE. For comparison purproses, we trained a VGG-FACE network using the training set of the AffectNet database (let us call this network 'the VGG-FACE baseline'). 

Table \ref{affectnet_expr_db} shows a comparison of the performance of: i) the VGG-FACE baseline, ii) the VGG-FACE network trained using the proposed approach and iii) AlexNet, the baseline network of the AffectNet database \cite{mollahosseini2017affectnet}.

\begin{table}[h]
\caption{AffectNet: Total accuracy and $F_1$ score of the VGG-FACE trained using the proposed approach vs state-of-the-art networks}
\label{affectnet_expr_db}
\centering
\scalebox{0.95}{
\begin{tabular}{ |c||c|c| }
 \hline
  Networks   & Total Accuracy & $F_1$ score  \\
 \hline
AlexNet \cite{mollahosseini2017affectnet}  & 0.58 & 0.58 	   \\
 \hline
 \begin{tabular}{@{}c@{}}the VGG-FACE baseline \end{tabular} & 0.52 & 0.51  	\\
 \hline 
\begin{tabular}{@{}c@{}}VGG-FACE trained \\ using the proposed approach \end{tabular} & \textbf{0.60} & \textbf{0.59}	 	\\
 \hline 
\end{tabular}
}
\end{table}

From Table \ref{affectnet_expr_db}, it can be verified that the network trained using the proposed approach outperformed all the other networks. In more detail, when compared to the VGG-FACE  baseline network, the boost in performance was significant, as also shown in Fig. \ref{conf_matrix_affectnet} in terms of the confusion matrices obtained by the two networks. This can be explained by the big size of the added synthesized images. When compared to the AffectNet's baseline, a slightly improved performance was also obtained; this could be higher, if we had synthesized images for the contempt category as well.

\noindent  \textbf{AFEW}. We synthesized 56,514  images from the AFEW database; this number was almost 1.4 times bigger than its training set size (41,406). The employed network architecture was VGG-FACE.
For comparison purposes, we first trained a baseline network on AFEW's training set, which we call the VGG-FACE baseline. For further comparisons, we used the following networks developed by the three winning methods of the EmotiW 2017 Grand Challenge: i) VGG-FACE-FER: the VGG-FACE was first  fine-tuned on the FER2013 database \cite{goodfellow2013challenges} and then trained on the AFEW as described in \cite{knyazev2017convolutional}, ii) VGG-FACE-external: the VGG-FACE was trained on the union of the AFEW database and some external data as described in \cite{vielzeuf2017temporal} and iii) VGG-FACE-LSTM-external-augmentation: the VGG-FACE-LSTM was trained on the union of the AFEW database and some external data; then data augmentation was performed, as described in \cite{vielzeuf2017temporal}.

\begin{table}[h]
\caption{AFEW: Total accuracy of the VGG-FACE trained using the proposed approach vs state-of-the-art networks}
\label{afew_db}
\centering
\scalebox{0.86}{
\begin{tabular}{ |c||c| }
 %\multicolumn{1}{|c||}{Networks} & \multicolumn{1}{|c}{Basic} \\
\hline
Networks   & Total Accuracy \\
\hhline{|=|=|} 
the VGG-FACE baseline  & 0.379		   \\
 \hline
 VGG-FACE-external \cite{vielzeuf2017temporal} & 0.414 \\
 \hline
  VGG-FACE-FER \cite{knyazev2017convolutional} & 0.483 \\
 \hline
  VGG-FACE-LSTM-external-augmentation \cite{vielzeuf2017temporal} & \textbf{0.486} \\
 \hline
\begin{tabular}{@{}c@{}}VGG-FACE trained \\ using the proposed approach \end{tabular} & 0.484 	\\
 \hline 
\end{tabular}
}
\end{table}

\begin{figure*}[h!]
\centering
\adjincludegraphics[height=6cm]{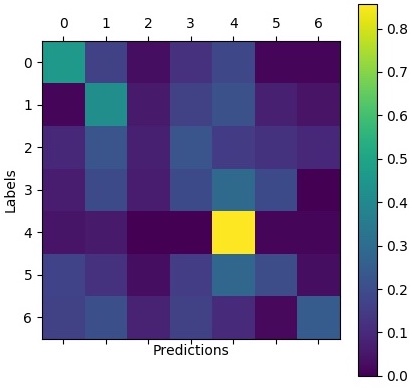}
 \HS \HS \HS \HS \HS 
 \adjincludegraphics[height=6.9cm,trim={0\totalheight} {.07\totalheight} {0\totalheight} {0\totalheight}]{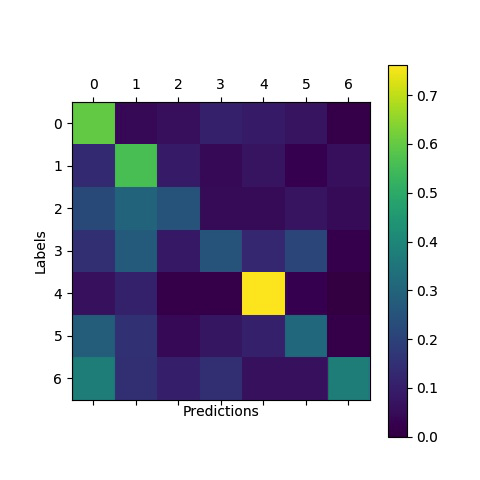}
 \\
  (a)\HS \HS \HS \HS \HS \HS \HS \HS \HS \HS \HS \HS \HS \HS \HS \HS \HS \HS \HS \HS \HS \HS \HS \HS \HS \HS \HS \HS \HS \HS \HS \HS \HS \HS \HS \HS \HS \HS \HS \HS \HS \HS \HS \HS \HS \HS \HS \HS \HS \HS \HS \HS \HS \HS \HS \HS \HS \HS \HS \HS \HS \HS \HS \HS \HS \HS \HS \HS \HS \HS \HS (b)\\
\caption{The confusion matrix of (a) the VGG-FACE baseline and (b) the VGG-FACE trained using the proposed approach for the AFEW database; 0: Neutral, 1: Anger, 2: Disgust, 3: Fear, 4: Joy, 5: Sadness, 6: Surprise}
\label{conf_matrix_afew}
\end{figure*}

Table \ref{afew_db} shows a comparison of the performance of the above described networks. From Table \ref{afew_db}, one can see that the VGG-FACE trained using the proposed approach performed much better than the same network trained on, either only the AFEW database, or the union of the AFEW database with some external data whose size in terms of videos was the same as that of AFEW. 
The boost in performance can be explained taking into account the fact that the fear, disgust and surprise classes contained few data in AFEW and that our approach augmented the data size of those classes; in total the large number of synthesized images assisted to improve the performance of the network.
This is evident when comparing the confusion matrix of the VGG-FACE baseline to the one of VGG-FACE trained using the proposed approach, as can be seen in  Fig.\ref{conf_matrix_afew}. The diagonal of the two confusion matrices indicates that there is an increase in the performance in almost all basic categories.

Additionally, performance of our network is slightly better than the performance of the same VGG-FACE network first fine-tuned on the FER2013 database and then trained on the AFEW. FER2013 is a database of around 35,000 still images and different identities, annotated with the six basic expressions. In this case, the network that was first fine-tuned on the FER2013 database has seen more faces, since the tasks were similar. However, still our network provided a slightly better performance. 
On the other hand, our network had a slightly worse performance than a VGG-FACE-LSTM  network that was trained with the same external data mentioned before and  was also trained with data augmentation. Here, it was the LSTM network, which due to the time recurrent nature could better exploit the fact that AFEW consists of video sequences.

\noindent  \textbf{BU-3DFE}. We synthesized 600  images from the BU-3DFE database. This number was almost one fourth of its size (2,500). BU-3DFE is a small database and is not really suited for training DNNs. 
The network architecture that we employed here was VGG-FACE, with a modification in the number of hidden units in the two first fully connected layers. Since we did not have a lot of data for training the network, we i) used 256 and 128 units in the two fully connected layers and ii) kept the convolutional weights fixed, training only the fully connected ones.
For training the network on this database, we used a 10-fold person-independent cross-validation strategy; in each fold, we augmented the training set with the synthesized images of people appearing only in that set (preserving person independence).
The reported total accuracy of the model has been the average of the total accuracies over the 10-folds.

At first, we trained the above described VGG-FACE network (let us call this network 'the VGG-FACE baseline'). Next, we trained the above described VGG-FACE network, but also applied on-the-fly data augmentation techniques, such as: small rotations, left and right flipping, first resize and then random crop to original dimensions, random brightness and saturation (let us call this network 'VGG-FACE-augmentation'). Finally, we trained  the above described VGG-FACE network using the proposed approach.

\begin{table}[h]
\caption{BU-3DFE: Total accuracy of the VGG-FACE trained using the proposed approach vs the VGG-FACE baseline and the VGG-FACE trained with on-the-fly data augmentation.}
\label{bu3dfe_db}
\centering
\begin{tabular}{ |c||c| }
 %\multicolumn{1}{|c||}{Networks} & \multicolumn{1}{|c}{Basic} \\
\hline
Networks   & Total Accuracy \\
\hhline{|=|=|} 
the VGG-FACE baseline  & 0.528		   \\
 \hline
 VGG-FACE-augmentation & 0.588 \\
 \hline
\begin{tabular}{@{}c@{}}VGG-FACE trained \\ using the proposed approach \end{tabular} & \textbf{0.768} 	\\
 \hline 
\end{tabular}
\end{table}

Table \ref{bu3dfe_db} shows a comparison of the performance of those networks.
From Table \ref{bu3dfe_db}, it can be verified that the network trained using the proposed approach greatly outperformed the networks trained without it. This indicates that the proposed approach for synthesizing images can be used for data augmentation in cases of small amount of DNN training data, being able to significantly improve the obtained performances.

%From both tables, it can be verified that the network trained on the augmented, with synthetic images, dataset, outperformed the networks trained without them. This implies that, by augmenting the original training set, our methodology improved the network performance. It should be noted that the boost in performance is greater when the number of augmented images is much greater  than the number of images in the dataset (which is the case of AFEW-VA that contains 30,000 frames, while the augmented set included 109,000 more frames).

\subsection{Quantitative evaluation of the facial affect synthesis used in testing or training tasks}

Results in the previous section show that the data generated using
our approach provide improvements in network performance in both valence-arousal and basic expressions settings, when used for data augmentation.
In the following, we perform further analysis (two different settings) to assess the quality of our generated data, compared to the data synthesized by StarGAN and Ganimation, focusing only on the synthesized data. 

In the first setting, the synthesized data are evaluated as a test set, for each database, against models trained on real data/images.
%\textcolor{red}{why we did this experiment? poio to intuition?}
%%% isws allaxe to baselines k episis prosthese to afew-va to kaneis me vasi ta alla models
%We evaluate the baseline models of Aff-Wild, RECOLA and AffectNet databases (described in Section \ref{va_augm}) on these test sets and report the performances. 
The AffWildNet that has been trained solely on Aff-Wild's training set, the ResNet-GRU trained on the RECOLA's training set and the VGG-FACE baseline trained on AffectNet's training set (all described in Section \ref{dimensional}), have been used as emotion regressors and are being evaluated on each of the three afore-mentioned synthesized datasets.  
From Table \ref{test_synth} it is evident that the networks trained on the afore mentioned databases displayed a much better performance (in all databases) when tested on the synthesized data from the proposed approach in comparison to the synthesized data from StarGAN and Ganimation. 

We further conducted a second setting, using the synthesized data to train respective DNN models. These models are then evaluated on the real test set of Aff-Wild, RECOLA and AffectNet. Table \ref{train_augm} shows the results of this setting.
The performance in terms of both CCC and MSE is much higher in all databases when the networks are trained with the data synthesized by the proposed approach. This difference in the compared performances, along with the former results, reflect the direct value of our generated data in enhancing regression performance.

\begin{table*}[htp]

\centering
\caption{CCC and MSE evaluation of valence \& arousal predictions provided by the: i) AffWildNet (trained on Aff-Wild), ii) ResNet-GRU (trained on RECOLA) and iii) the VGG-FACE baseline (trained on AffectNet); these networks are tested on the synthesized images by StarGAN, Ganimation and our approach. Each score is shown in the format: Valence value-Arousal value}
\label{test_synth}
\begin{tabular}{ c||c|c|c|c|c}
Databases      & Methods  & Evaluation Metrics &  \multicolumn{3}{c}{Networks} \\
\hline
& & &  AffWildNet\cite{kollias2019deep} & ResNet-GRU\cite{kollias2019deep}  &  the VGG-FACE baseline \\
\hhline{=:=:=:=:=:=}
 \multirow{3}{*}{Aff-Wild} &StarGAN & \begin{tabular}{@{}c@{}} CCC \\ MSE \end{tabular} &\begin{tabular}{@{}c@{}} 0.33-0.26 \\ 0.21-0.19 \end{tabular}   &-  &-\\
  \cline{2-6}
    &Ganimation& \begin{tabular}{@{}c@{}} CCC \\ MSE \end{tabular} & \begin{tabular}{@{}c@{}}0.35-0.28  \\ 0.19-0.16 \end{tabular} & - &-\\
     \cline{2-6}
    & Ours & \begin{tabular}{@{}c@{}} CCC \\ MSE \end{tabular} & \begin{tabular}{@{}c@{}} \textbf{0.43-0.33}  \\ \textbf{0.15-0.13} \end{tabular} & - & -\\
\hline
\hline
 \multirow{3}{*}{RECOLA} &StarGAN & \begin{tabular}{@{}c@{}} CCC  \end{tabular} & - & \begin{tabular}{@{}c@{}} 0.29-0.23 \end{tabular} &-\\
  \cline{2-6}
    &Ganimation& \begin{tabular}{@{}c@{}} CCC  \end{tabular} & -  & \begin{tabular}{@{}c@{}} 0.28-0.22 \end{tabular} & - \\
     \cline{2-6}
    & Ours & \begin{tabular}{@{}c@{}} CCC  \end{tabular} &  -& \begin{tabular}{@{}c@{}} \textbf{0.34-0.33} \end{tabular} &- \\
\hline
\hline
 \multirow{3}{*}{AffectNet} &StarGAN & \begin{tabular}{@{}c@{}} CCC \\ MSE \end{tabular} & - & - & \begin{tabular}{@{}c@{}}  0.23-0.23 \\ 0.34-0.37  \end{tabular} \\
  \cline{2-6}
    &Ganimation& \begin{tabular}{@{}c@{}} CCC \\ MSE \end{tabular} & - & - & \begin{tabular}{@{}c@{}} 0.26-0.21 \\   0.31-0.38 \end{tabular} \\
     \cline{2-6}
    & Ours & \begin{tabular}{@{}c@{}} CCC \\ MSE \end{tabular} & - & - &  \begin{tabular}{@{}c@{}} \textbf{0.39-0.31} \\ \textbf{0.27-0.28}  \end{tabular} \\
\hline
\end{tabular}
\end{table*}

\begin{table*}[htp]

\centering
\caption{CCC and MSE evaluation of valence \& arousal predictions provided by the: i) AffWildNet, ii) ResNet-GRU and iii) the VGG-FACE baseline; these networks are trained on the synthesized images by StarGAN, Ganimation and our approach; these networks are evaluated on the Aff-Wild, RECOLA and AffectNet test sets. Each score is shown in the format: Valence value-Arousal value}
\label{train_augm}
\begin{tabular}{ c||c|c|c|c|c}
Databases      & Methods  & Evaluation Metrics &  \multicolumn{3}{c}{Networks} \\
\hline
& & &  AffWildNet & ResNet-GRU  &  VGG-FACE baseline \\
\hhline{=:=:=:=:=:=}
 \multirow{3}{*}{Aff-Wild} &StarGAN & \begin{tabular}{@{}c@{}} CCC \\ MSE \end{tabular} & \begin{tabular}{@{}c@{}} 0.16-0.13 \\ 0.18-0.17 \end{tabular} & - & - \\
  \cline{2-6}
    &Ganimation& \begin{tabular}{@{}c@{}} CCC \\ MSE \end{tabular} & \begin{tabular}{@{}c@{}} 0.17-0.14 \\ 0.17-0.15 \end{tabular}  & - & - \\
     \cline{2-6}
    & Ours & \begin{tabular}{@{}c@{}} CCC \\ MSE \end{tabular} & \begin{tabular}{@{}c@{}} \textbf{0.21-0.20} \\ \textbf{0.15-0.12} \end{tabular}   & - & - \\
\hline
\hline
 \multirow{3}{*}{RECOLA} &StarGAN & \begin{tabular}{@{}c@{}} CCC \end{tabular} & - & 0.19-0.10  & - \\
  \cline{2-6}
    &Ganimation& \begin{tabular}{@{}c@{}} CCC  \end{tabular} & -  & 0.17-0.10  & - \\
     \cline{2-6}
    & Ours & \begin{tabular}{@{}c@{}} CCC \end{tabular} & - & \textbf{0.23-0.14} & - \\
\hline
\hline
 \multirow{3}{*}{AffectNet} &StarGAN & \begin{tabular}{@{}c@{}} CCC \\ MSE \end{tabular} & -  & -  & \begin{tabular}{@{}c@{}} 0.37-0.29 \\ 0.23-0.21 \end{tabular}  \\
  \cline{2-6}
    &Ganimation& \begin{tabular}{@{}c@{}} CCC \\ MSE \end{tabular} & - & - & \begin{tabular}{@{}c@{}} 0.40-0.31 \\ 0.20-0.19 \end{tabular}   \\
     \cline{2-6}
    & Ours & \begin{tabular}{@{}c@{}} CCC \\ MSE \end{tabular} & - & - & \begin{tabular}{@{}c@{}} \textbf{0.45-0.35} \\ \textbf{0.18-0.17} \end{tabular}  \\
\hline
\end{tabular}
\end{table*}

%% gia granularity grapsto
%Results in the previous section show that the data generated using cGAN models provide improvements in classification performance in low resource settings. I perform further analysis to assess the impact of data size in training cGAN models, as well as a visual assessment of the generated fake data. Using the twitter dataset, I perform two sets of analysis: (i) evaluating the impact of the training dataset size on the cGAN model.  I discuss them in detail below

%% gia granularity grapsto I evaluate the impact of the size of the available dataset on the cGAN model in this section. The quality of the cGAN data generated is evaluated in two settings: when used as a training set as well as a test set. The setup of my analysis is as follows.

\subsection{Effect of synthesized data granularity on performance improvement}

\begin{figure*}[h!]
\centering
\adjincludegraphics[height=4.3cm]{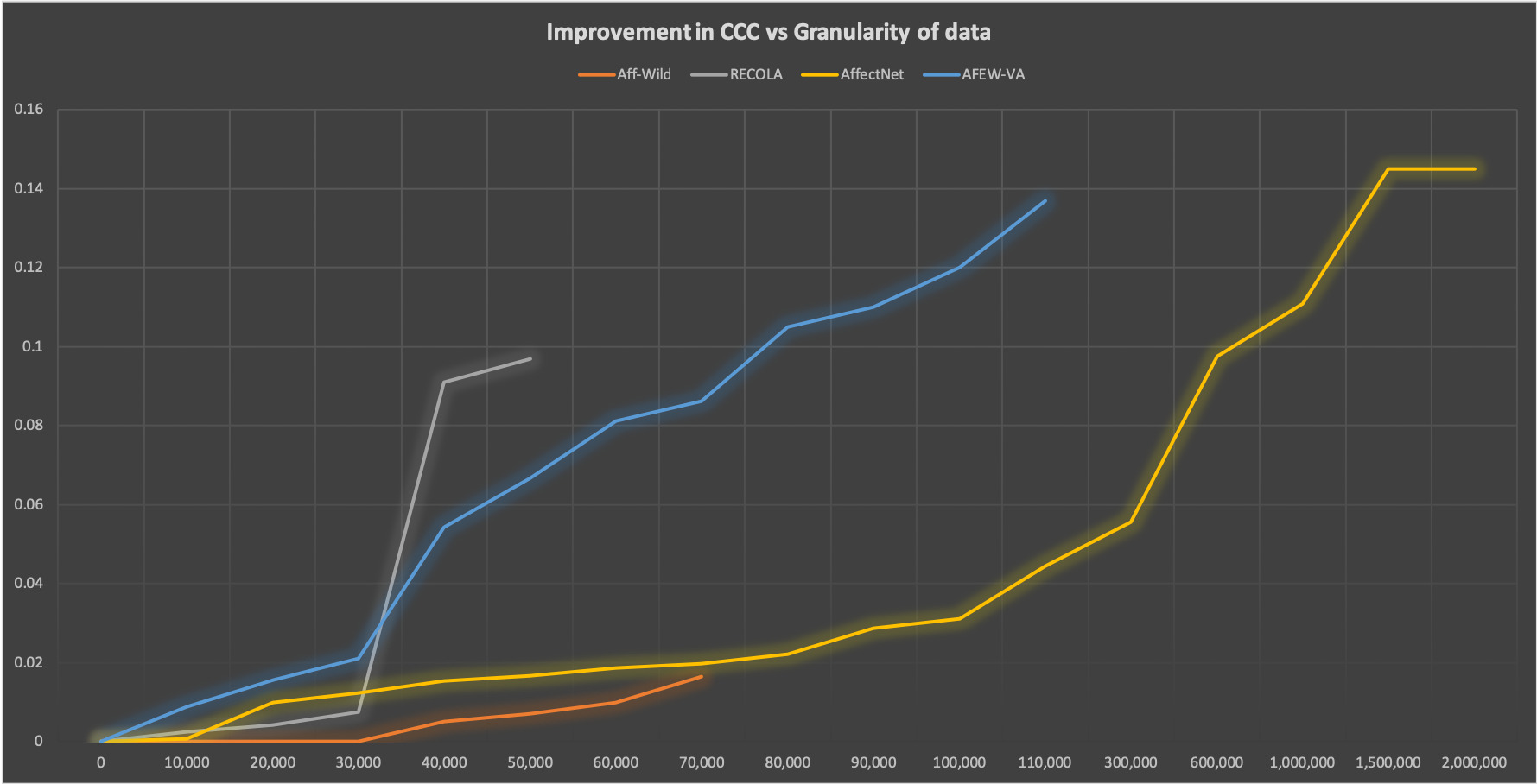}
\adjincludegraphics[height=4.3cm]{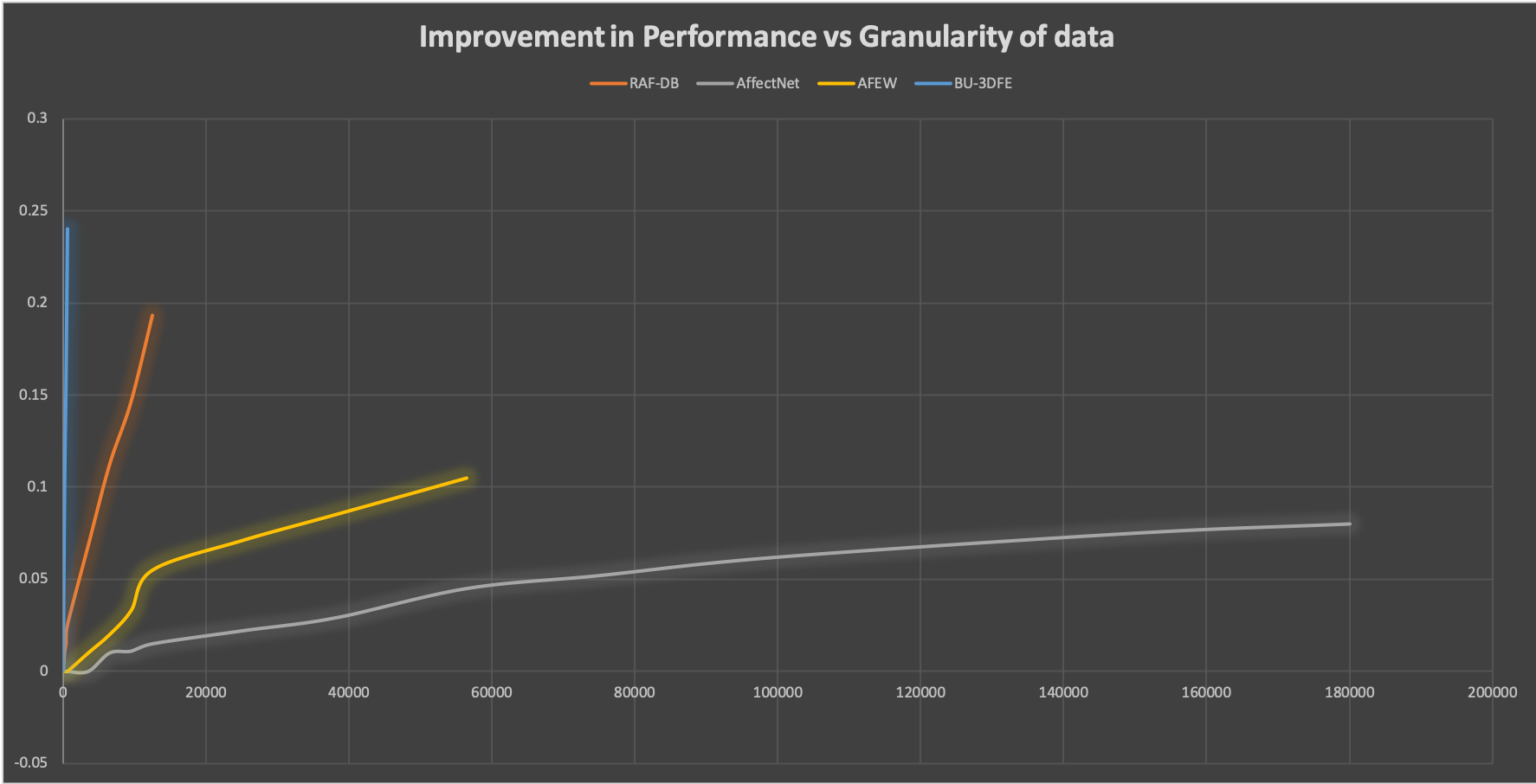}\\
(a)\HS \HS \HS \HS \HS \HS \HS \HS \HS \HS \HS \HS \HS \HS \HS \HS \HS \HS \HS \HS \HS \HS \HS \HS \HS \HS \HS \HS \HS \HS \HS \HS \HS \HS \HS \HS \HS \HS \HS \HS \HS \HS \HS \HS \HS \HS \HS \HS \HS \HS \HS \HS \HS \HS \HS \HS \HS \HS \HS \HS \HS \HS \HS \HS \HS \HS \HS \HS \HS \HS \HS \HS \HS \HS \HS \HS \HS \HS \HS \HS \HS \HS (b)\\
\caption{Improvement in network performance vs amount of synthesized data; criteria: (a) mean/average CCC of VA in Aff-Wild, RECOLA, AffectNet, AFEW-VA and (b) mean diagonal value of the confusion matrix for RAF-DB, F1 score for AffectNet,  Total Accuracy for AFEW and BU-3DFE }
\label{improve_amount}
\end{figure*}

In this subsection we performed experiments using a subset of our synthesized data for augmenting the data-bases. Our aim is to see if all synthesized data are needed for augmenting network training and more generally to see how the improvement in classification and regression scale with the granularity of synthesized data.
In more detail, for each database used in our experiments, we used a subset of $N$ synthesised data from this database to augment its training set. Table \ref{subset_gran} shows the databases and its corresponding $N$ values.

\begin{table*}
\centering
\caption{Databases used in our approach and the different values of $N$ for each one; $N$ denotes a subset of the synthesized data (per database) by the proposed approach}
\label{subset_gran}
\begin{tabular}{c|c}
Databases & $N$ synthesized data  \\
\hhline{-|-} 
\hhline{-|-} 
\hhline{-|-} 
\hhline{-|-} 
Aff-Wild &  $ N \in \{10K, 20K, 30K, 40K, 50K, 60K$\} \\
\hline
RECOLA &  $ N \in \{10K, 20K, 30K, 40K, 50K$\} \\
\hline
AffectNet (VA) &  \begin{tabular}{@{}c@{}} $ N \in \{10K, 20K, 30K, 40K, 50K, 60K, 70K, 80K, 90K, 100K, 110K, 300K, 600K,$ \\  $1M, 1.5M, 2M, 2.5M$\} \end{tabular} \\
\hline
AFEW-VA &  $ N \in \{10K, 20K, 30K, 40K, 50K, 60K, 70K, 80K, 90K, 100K, 110K$\} \\
\hhline{=|=} 
RAF-DB &  $ N \in \{200, 400, 600, 3.5K, 6.5K, 9.5K, 12.5K$\} \\
\hline
AffectNet (Expressions) & $ N \in \{6.5K, 12.5K, 25K, 38K, 56.5K, 75K, 100K, 150K, 180K$\} \\ 
\hline
AFEW &  $ N \in \{3.5K, 6.5K, 12.5K, 25K, 38K, 56.5K$\} \\
\hline
BU-3DFE &  $ N \in \{200, 400, 600$\} \\
\hline
\end{tabular}
\end{table*}

Fig. \ref{improve_amount} shows the improvement in network  performance when training using additionally auxiliary data; the improvement shown per database is the difference in the performances when training networks with only the database's training set and when training them with the union of the training set and auxiliary data. Fig. \ref{improve_amount} illustrates for each database the difference in network performance, when $N$ synthesized data generated by our approach ($N$ defined in  Table \ref{subset_gran}) are used as auxiliary data.

The performance measure for Aff-Wild, RECOLA, AffectNet and AFEW-VA is the average of valence CCC and arousal CCC. The performance measure for the rest databases depends on the database. More details follow.

\bigskip

\noindent \textbf{Dimensional affect generation} 

\bigskip

\noindent For the Aff-Wild database, we use the VGG-FACE-GRU network. 
%and trained it with the union of Aff-Wild's training set and synthesized images from our approach. 
When augmenting the dataset with $30K$ or less synthesized images, no performance improvement is seen, whereas when augmenting it with more than $30K$, the performance is increasing, following the increase in the granularity of synthesized data. Adding synthesized data to the training set seems to be beneficial for improving the performance and thus the improvement would be much greater if we added more than $60K$ (if we had more neutral expressions), although probably at a given  point,  a  plateau  would  be reached (considering the large training set that consists of around $1M$ images).
%Nevertheless, the total observed improvement is not a lot because the total (maximum) number of synthesized images was small compared to the training set. 

For the RECOLA database, we use the ResNet-GRU network. 
%and trained it with the union of RECOLA's training set and synthesized images from our approach. 
When augmenting the dataset with up to $30K$ synthesized images, there exists small performance improvement, whereas when augmenting it with more than $30K$, the performance is continuously increasing following the increase in the granularity of synthesized data; this increase is large. This is expected, since $120K$ frames are not sufficient for training a network for regression and additionally, $170K$ frames are not either.

For the AffectNet database, we use the VGG-FACE network.
%and trained it with the union of AffectNet's training set and synthesized images from our approach. 
After adding $10K$ synthesized images, the performance starts to increase. This increase continues to happen as more data are added until the training set has been augmented with $1.5M$ data. If more data are added, the performance does not change, implying that a plateau has been reached. The final performance improvement is large. 

For the AFEW-VA database, we use the VGG-FACE-GRU network.
%and trained it with the union of AFEW-VA's training set and synthesized images from our approach. 
The improvement is systematically very significant. When adding more than $30K$ data, the increase in performance is more rapid. The performance is expected to continue increasing while more data are added, as both the initial training set of around $23K$ frames and the augmented set of around $135K$ frames are not large enough to train a DNN for regression.

\bigskip

\noindent \textbf{Categorical affect generation}

\bigskip

\noindent For the RAF-DB database, we use the VGG-FACE network and the performance is measured in terms of the mean diagonal value of the confusion matrix. The increase in performance is almost linear as more data are used. The final performance gain is great. RAF-DB is a very small database (of size about $12K$ images) and therefore if we had more data to add, the performance would further improve.

In the AffectNet database, we use the VGG-FACE network and performance is measured in terms of the F1 score. Increasing the amount of added data provides a respective increase in the performance. After adding $60K$ images the performance is increasing at a lower rate. It should be mentioned that the results include erroneous classification of the contempt class. If we synthesized samples of the contempt class as well, the network would provide a higher performance; but this is beyond the scope of the current paper. 

In the AFEW database, we use the VGG-FACE network; the performance measure is total accuracy. The performance is increasing with the addition of more data. The performance increase is significant. The AFEW database is a small database (of size about $40K$ images) and therefore adding data is expected to increment the performance.

In the BU-3DFE database, we use the VGG-FACE network; the performance measure is total accuracy. There is a huge and rapid increase in network performance with the addition of data.
This is explained by the very small size of BU-3DFE (around $2K$) which makes it impossible to train a neural network on it.

General deductions that can be made from Fig. \ref{improve_amount}:
\begin{itemize}[leftmargin=*,noitemsep,nolistsep]
\item the smaller the size of the database, the bigger and faster the increase in performance would be, when augmenting it with synthesized data from our approach 
\item the improvement in performance is small if we augment the training set with few data in proportion to its size 
\item in dimensionally annotated databases, a plateau is reached and no further improvement is seen when a lot of data (about $\geq 1.5M$ in our case) are added
\item  the performance due to data augmentation does not increase commensurately; in the AffectNet database (mainly in the valence-arousal case) the gain yielded by data augmentation saturates as N increases
\item generally, the performance increase is larger in categorically annotated databases in comparison to dimensionally annotated ones. This is an interesting result, since it indicates that synthesizing more data is needed in the latter case, to make the data distribution more dense.
\end{itemize}

%By observing the two tables, some useful facts emerge. First, cross-dataset learning seems to be beneficial in general, except for the single-shot modality when testing on the CUHK03 dataset, where the performance essentially does not change. Notably, fine-tuning gives better results when it is carried out with Market-1501 on the network trained from scratch on CUHK03, than vice-versa

%This is an interesting result, since it indicates that, other than being an economic and effective proxy for real data, the SOMA framework appears to produce a nice general optimization of the network, that later can be properly specialized using the data where the classifier will be applied.

%As one would expect,

%Certainly, at a given point,a plateau should be reached, and finding this point is a key open issue.

\subsection{Effect of subjects' age in classification \& regression results}

It is interesting to quantitatively assess the effect of age on the performance of the proposed approach. However, not all databases contain age information about their subjects. To achieve this, we trained an age estimator on them. In more detail, we trained a Wide Residual Network (WideResNet) \cite{zagoruyko2016wide} on the union of IMDB \cite{rothe2015dex} and Adience datasets \cite{eidinger2014age} (so that the training dataset contained an adequate number of images of people under the age of 25) and tested it on WIKI \cite{rothe2015dex}. Then we applied this estimator on the test sets of the examined databases.

\begin{table*}[h]
\caption{Age Analysis in terms of CCC and MSE for the dimensionally annotated databases}
\label{age_va}
\centering
%\scalebox{0.86}{
\begin{tabular}{ c|c|c|c|c|c|c|c }
\multicolumn{1}{c|}{Databases}   & \multicolumn{1}{c|}{Ages} & \multicolumn{1}{c|}{\# Test Samples} & \multicolumn{1}{c|}{\# Synthesized Samples} & \multicolumn{2}{c|}{Network-Augmented} & \multicolumn{2}{c}{Network}\\
\hline 
& & & & CCC & MSE & CCC & MSE \\
\hhline{=|=|=|=|=|=|=|=}
Aff-Wild 
& \begin{tabular}{@{}c@{}} 20-29 \\ 30-39 \\ 40-49 \\ 50-59  \end{tabular}
& \begin{tabular}{@{}c@{}} 29,013 \\ 99,962 \\ 44,727 \\ 41,748  \end{tabular}
& \begin{tabular}{@{}c@{}} 5,301 \\ 23,427 \\ 21,831 \\ 9,120  \end{tabular}
& \begin{tabular}{@{}c@{}} \textbf{0.61-0.38} \\ \textbf{0.66-0.47} \\ \textbf{0.50-0.48} \\ \textbf{0.58-0.40} \end{tabular} 
& \begin{tabular}{@{}c@{}} \textbf{0.101-0.063} \\ \textbf{0.077-0.054} \\ \textbf{0.048-0.033} \\ \textbf{0.074-0.054} \end{tabular} 
& \begin{tabular}{@{}c@{}} 0.59-0.37 \\ 0.61-0.44 \\ 0.46-0.44 \\ 0.57-0.38  \end{tabular} 
& \begin{tabular}{@{}c@{}} 0.102-0.066 \\ 0.088-0.066 \\ 0.054-0.044 \\ 0.075-0.057  \end{tabular}    \\
\cline{2-8}
& total & 215,450 & 59,679 & \textbf{0.60-0.45} & \textbf{0.074-0.051} & 0.57-0.43 & 0.080-0.060\\ 
\hline
\hline
RECOLA 
& \begin{tabular}{@{}c@{}} 30-39 \\ 40-49 \\ 50-59 \end{tabular} 
& \begin{tabular}{@{}c@{}}  90,000 \\ 15,000 \\ 7,500 \end{tabular}
& \begin{tabular}{@{}c@{}}  11,001  \\ 16,188 \\ 11,742 \end{tabular}
& \begin{tabular}{@{}c@{}}  \textbf{0.61-0.38} \\ \textbf{0.43-0.24} \\ \textbf{0.49-0.20} \\  \end{tabular} 
& \begin{tabular}{@{}c@{}} - \\ - \\ -  \end{tabular} 
& \begin{tabular}{@{}c@{}}  0.60-0.34 \\ 0.36-0.19 \\ 0.44-0.10 \\  \end{tabular} 
& \begin{tabular}{@{}c@{}} - \\ - \\ -  \end{tabular} \\
\cline{2-8}
& total & 112,500 & 38,931 & \textbf{0.55-0.31} & - & 0.53-0.27 & -\\ 
\hline
\hline
AffectNet 
& \begin{tabular}{@{}c@{}}  0-19 \\ 20-29 \\ 30-39 \\ 40-49 \\ 50-59 \\ 60-89  \end{tabular} 
& \begin{tabular}{@{}c@{}}   172 \\ 1,179 \\ 1,218 \\ 762 \\ 569 \\ 600   \end{tabular}
& \begin{tabular}{@{}c@{}}  118,902 \\ 714,232 \\ 814,588 \\ 452,504 \\ 229,938 \\ 146,091    \end{tabular}
& \begin{tabular}{@{}c@{}}   \textbf{0.67-0.55} \\ \textbf{0.60-0.53} \\ \textbf{0.64-0.54} \\ \textbf{0.64-0.61}  \\ \textbf{0.58-0.53}  \\  \textbf{0.62-0.44}  \end{tabular} 
& \begin{tabular}{@{}c@{}}  \textbf{0.105-0.156} \\ \textbf{0.128-0.159} \\ \textbf{0.139-0.145}  \\ \textbf{0.149-0.134} \\ \textbf{0.161-0.149} \\ \textbf{0.145-0.167}  \end{tabular} 
& \begin{tabular}{@{}c@{}}  0.61-0.41 \\ 0.51-0.36 \\ 0.50-0.39 \\ 0.49-0.44 \\ 0.47-0.34 \\ 0.51-0.29 \end{tabular} 
& \begin{tabular}{@{}c@{}}  0.127-0.181 \\ 0.170-0.193 \\ 0.193-0.169  \\ 0.202-0.166  \\ 0.216-0.181  \\ 0.200-0.195     \end{tabular} 
   \\
\cline{2-8}
& total & 4,500 & 2,476,235 & \textbf{0.62-0.54} & \textbf{0.141-0.150} & 0.50-0.37 & 0.190-0.180\\ 
\hline
\hline
AFEW-VA 
& \begin{tabular}{@{}c@{}} 20-29 \\ 30-39 \\ 40-49 \\ 50-59 \\ 60-79   \end{tabular} 
& \begin{tabular}{@{}c@{}}  766 \\ 1,990 \\ 1,558 \\ 946 \\ 396 \end{tabular}
& \begin{tabular}{@{}c@{}}  17,466  \\ 36,388 \\ 34,906 \\ 15,102 \\ 4,102  \end{tabular}
& \begin{tabular}{@{}c@{}}   0.46-0.60 \\ 0.51-0.62 \\ 0.59-0.47  \\ 0.74-0.85  \\  0.63-0.45   \end{tabular} 
& \begin{tabular}{@{}c@{}}  0.192-0.084  \\ 0.254-0.080 \\ 0.211-0.076 \\ 0.215-0.045 \\ 0.236-0.100   \end{tabular} 
& \begin{tabular}{@{}c@{}}   - \\- \\ - \\ -  \\ -      \end{tabular}
& \begin{tabular}{@{}c@{}} - \\- \\ - \\ -  \\ -       \end{tabular}
\\
\cline{2-8}
& total & 5,646 & 108,864 & 0.57-0.59 & 0.226-0.075 & - & -\\ 
\hline
\end{tabular}
%}
\end{table*}

\begin{table*}[h]
\caption{Age Analysis for the categorically annotated databases; criterion for RAF-DB \& AffectNet is F1 score, for AFEW \& BU-3DFE is total accuracy; AFEW test samples refer to: number of videos (frames)}
\label{age_expr}
\centering
%\scalebox{0.86}{
\begin{tabular}{ c|c|c|c|c|c }
\multicolumn{1}{c|}{Databases}   & \multicolumn{1}{c|}{Ages} & \multicolumn{1}{c|}{\# Test Samples} & \multicolumn{1}{c|}{\# Synthesized Samples} & \multicolumn{1}{c|}{VGG-FACE-Augmented} & \multicolumn{1}{c}{VGG-FACE}\\
\hline
& & & & Performance Metric & Performance Metric \\
\hhline{=|=|=|=|=|=}
RAF-DB
& \begin{tabular}{@{}c@{}} 10-19 \\ 20-29 \\ 30-39 \\ 40-49 \\ 50-59 \\ 60-69 \\ 70-79 \end{tabular}
& \begin{tabular}{@{}c@{}} 168 \\ 911 \\ 998 \\ 516 \\ 258 \\ 149 \\ 68  \end{tabular}
& \begin{tabular}{@{}c@{}} 210 \\ 2,250  \\ 4,320 \\ 3,606  \\ 1,776 \\   552 \\ 128   \end{tabular}
& \begin{tabular}{@{}c@{}} \textbf{0.631} \\ \textbf{0.813} \\ \textbf{0.739} \\ \textbf{0.744} \\ \textbf{0.709} \\ \textbf{0.657} \\ \textbf{0.904} \end{tabular} 
& \begin{tabular}{@{}c@{}} 0.446 \\ 0.556 \\ 0.498 \\ 0.511 \\ 0.440 \\ 0.550 \\ 0.635  \end{tabular} 
\\
\cline{2-6}
& total & 3,068 & 12,828 & \textbf{0.738} & 0.505 \\ 
\hline
\hline
AffectNet 
& \begin{tabular}{@{}c@{}} 0-19 \\ 20-29 \\ 30-39 \\ 40-49 \\ 50-59 \\ 60-69 \\ 70-79 \\ 80-89  \end{tabular} 
& \begin{tabular}{@{}c@{}}  152 \\ 882 \\ 962 \\ 594 \\ 431 \\ 289 \\ 161 \\ 29   \end{tabular}
%original one & \begin{tabular}{@{}c@{}} 0  \\ 13,584  \\ 76,710 \\ 54,390  \\ 20,226 \\ 7,908 \\ 2,874  \\ 534 \\  108    \end{tabular}
& \begin{tabular}{@{}c@{}} 12,516  \\ 45,182 \\ 55,513  \\ 27,632 \\ 20,204 \\ 11,178  \\ 3,582 \\  618    \end{tabular}
& \begin{tabular}{@{}c@{}}  \textbf{0.593} \\ \textbf{0.584} \\ \textbf{0.593} \\ \textbf{0.586} \\ \textbf{0.648} \\ \textbf{0.564} \\ \textbf{0.466} \\ \textbf{0.448} \end{tabular} 
& \begin{tabular}{@{}c@{}}  0.453 \\ 0.477\\ 0.518 \\ 0.532 \\ 0.606 \\ 0.498 \\ 0.398 \\ 0.410 \end{tabular} 
   \\
\cline{2-6}
& total & 3,500 & 176,425 & \textbf{0.590} & 0.510 \\ 
\hline
\hline
AFEW
& \begin{tabular}{@{}c@{}} 20-29 \\ 30-39 \\ 40-49 \\ 50-59 \\ 60-79  \end{tabular} 
& \begin{tabular}{@{}c@{}}  29 (1,536) \\ 156 (8,568) \\ 132 (7,803) \\ 57 (3,202) \\ 16 (764)  \end{tabular}
& \begin{tabular}{@{}c@{}}  6,474  \\ 22,518  \\ 17,934  \\ 7,482  \\ 2,106   \end{tabular}
& \begin{tabular}{@{}c@{}}  \textbf{0.379}  \\ \textbf{0.455} \\ \textbf{0.553} \\ \textbf{0.474} \\ \textbf{0.438}    \end{tabular}
& \begin{tabular}{@{}c@{}}  0.241  \\ 0.333 \\ 0.439  \\ 0.456 \\ 0.313    \end{tabular}
\\
\cline{2-6}
& total & 390 (21,873) & 56,514 & \textbf{0.484} & 0.379 \\ 
\hline
\hline
BU-3DFE
& \begin{tabular}{@{}c@{}} 20-29 \\ 30-39 \\ 40-49 \\ 50-59 \\ 60-70    \end{tabular} 
& \begin{tabular}{@{}c@{}}  115  \\ 100 \\ 100  \\ 100  \\ 85    \end{tabular}
& \begin{tabular}{@{}c@{}}  192  \\ 240  \\ 120  \\ 30   \\ 18   \end{tabular}
& \begin{tabular}{@{}c@{}}  \textbf{0.800}  \\ \textbf{0.820} \\ \textbf{0.800}  \\ \textbf{0.790}  \\  \textbf{0.600}   \end{tabular}
& \begin{tabular}{@{}c@{}}  0.600  \\ 0.570 \\ 0.550  \\ 0.490  \\  0.400   \end{tabular}
\\
\cline{2-6}
& total & 500 & 600 & \textbf{0.768} & 0.528 \\ 
\hline
\end{tabular}
%}
\end{table*}

Table \ref{age_va} shows, for each dimensionally annotated database (Aff-Wild, RECOLA, AffectNet and AFEW-VA), the estimated age groups (we split the age values into appropriate groups so that each group contained a significant amount of samples), the number of test samples that are within the age groups, the number of synthesized by our approach samples for each age group, different evaluation metrics (CCC and MSE) for each age group in two cases: when a network trained only with the training set of each database was used (denoted as 'Network' in Table \ref{age_va}) and when the same network was trained with the training set augmented with our approach's synthesized data (denoted as 'Network-Augmented' in Table \ref{age_va}). For  Aff-Wild and AFEW-VA, the VGG-FACE-GRU network was used, for RECOLA the ResNet-GRU and for AffectNet the VGG-FACE. 
%It is evident from Table \ref{age_va} that in all age groups

Table \ref{age_expr} is similar to Table \ref{age_va} with the difference being that it refers to categorically annotated databases (RAF-DB, AffectNet, AFEW and BU-3DFE). In this case, the evaluation metrics are the F1 score for RAF-DB and AffectNet, and the total accuracy for AFEW and BU-3DFE. The 'VGG-FACE-Augmented' refers to the case in which the VGG-FACE network is trained on the union of training set of each database and data synthesized by our approach.

By observing the two Tables (\ref{age_va} and \ref{age_expr}), it is seen that augmenting the training dataset with the images generated by our approach is beneficial in all age groups, both for regression and classification. It would be interesting to focus on specific groups, such as very young ($<$20 years old) in RAF-DB and AffectNet, each containing more than 150 subjects, or elderly (e.g., 70-79 years old) in AffectNet, also containing more than 150 subjects. In the former case, the F1 value improved from about 0.45 to 0.6; the F1 values over all categories improved from about 0.51 to 0.66. Although the F1 values in the very young category were lower than the mean F1 values over all ages, the improvement in both cases was similar. A similar observation can be made in the latter case, of elderly persons, with the F1 value in the category being improved from about 0.4 to 0.47. Although these values were lower than the total F1 values over all ages, which were 0.51 and 0.59 respectively, the improvement in these cases was similar as well.
This verifies the above-mentioned observation that the proposed approach for data augmentation can be also beneficial in cases where the number of available samples is rather small.

%Notably, fine-tuning gives better results when it is carried out with Market-1501 on the network trained from scratch on CUHK03, than vice-versa

% Moreover, we show clear benefit from finetuning on real data after training on synthetic data

%Using synthetic data often requires the additional step of validating against real-world data

%%%% grapse gia emotiw oti accuracy per video; gia affectnet pws oxi test set ara split to train se train+valid k use to valid as test
%%%% add results emotiw new network 0.379; rafdb new net f1 50.5 mean diag kapou 60
%%%% check me ta sequences pws tha to kaneis

\section{Conclusions and Future Work}

A novel approach to generate facial affect in faces has been presented in this paper. It leverages a dimensional emotion model in terms of valence and arousal or the six basic expressions, and a large scale 4D face database, the 4DFAB. We performed dimensional annotation of the 4DFAB and used the facial images with their respective annotations to generate mean faces on a discretized 2-D affect space. 

A methodology has been proposed  using these mean faces to synthesize faces with affect, both categorical or dimensional, static or dynamic. 
Using a given neutral image and the desired affect, which can be a Valence Arousal pair of values, a path in the 2D VA space, or one of the basic expression categories, the proposed approach performs face detection and landmark localization on the input neutral image, fits  a 3D Morphable Model on the resulting image, deforms the reconstructed face, adds the input affect and blends the new face with the given affect into the original image. 

An extensive experimental study has been conducted, providing both qualitative and quantitative evaluations of the proposed approach. The qualitative results show the achieved higher quality of the synthesized data compared to GAN-generated facial affect. The quantitative results are based on using the synthesized facial images for data augmentation and training of Deep Neural Networks over eight databases, annotated with either dimensional or categorical  affect  labels. It has been  shown  that, over all databases,  the  achieved  performance is much higher than i) the performance of the respective state-of-the-art methods, ii) the performance of the same DNNs with data augmentation provided by the StarGAN and Ganimation networks.

In our future work we will extend this approach to synthesize, not only dimensional, as well as categorical, affect in faces, but also Facial Action Units. In this way a Global Local synthesis of facial affect will be possible, through a unified modeling of global dimensional emotion and local action unit based facial expression synthesis. Another future direction will be to generate faces of different genders and human races.

%%%% svise ton teke 
\begin{acknowledgements}
The work of Stefanos Zafeiriou was partially funded by the FiDiPro program of Tekes with project number 1849/31/2015. The works of Dimitrios Kollias, as well as Evangelos Ververas were funded by Teaching Fellowships of Imperial College London. We also
thank the NVIDIA Corporation for donating a Titan X GPU. Additionally, we would like to thank the reviewers for their  valuable comments that helped us to improve this paper.
\end{acknowledgements}

\bibliographystyle{spmpsci}    

\bibliography{egbib}   % name your BibTeX data base

\end{document}